\documentclass[journal]{IEEEtran}

\usepackage{hyperref}
\usepackage{graphicx}
\usepackage{subfigure}
\usepackage[flushleft]{threeparttable}
\usepackage{booktabs} 
\usepackage{amsfonts}
\usepackage{multirow}
\usepackage{tabularx}
\usepackage{xcolor}
\usepackage{footnote}
\usepackage{caption}
\usepackage{textcomp, gensymb}
\usepackage{amsmath}

\usepackage{algorithm}
\usepackage{algpseudocode}

\newcolumntype{M}{>{$}c<{$}}
\newcolumntype{Z}{>{\centering\arraybackslash}X}
\newcolumntype{C}{>{\centering\arraybackslash}p}
\newcolumntype{Y}{>{\centering\arraybackslash}X}
\newcolumntype{L}{>{\raggedright\arraybackslash}p}

\newcommand{\specialcell}[2][c]{%
  \begin{tabular}[#1]{@{}c@{}}#2\end{tabular}}

\newcommand{\leftcell}[2][l]{%
  \begin{tabular}[#1]{@{}l@{}}#2\end{tabular}}

\ifCLASSINFOpdf
\else
\fi

\begin{document}
\title{Latent Fingerprint Recognition: Fusion of Local and Global Embeddings}

\author{Steven~A.~Grosz~and~Anil~K.~Jain,~\IEEEmembership{Life~Fellow,~IEEE}
\thanks{S.A. Grosz and A.K. Jain are with the Department of Computer Science and Engineering, Michigan State University, East Lansing, MI, 48824 USA (e-mail: groszste@cse.msu.edu, jain@cse.msu.edu).}
}

\markboth{Journal of \LaTeX\ Class Files,~Vol.~14, No.~8, August~2015}%
{Grosz and Jain: Latent Fingerprint Recognition: Fusion of Local and Global Embeddings}

\maketitle

\begin{figure}[ht!]
\includegraphics[width=\linewidth]{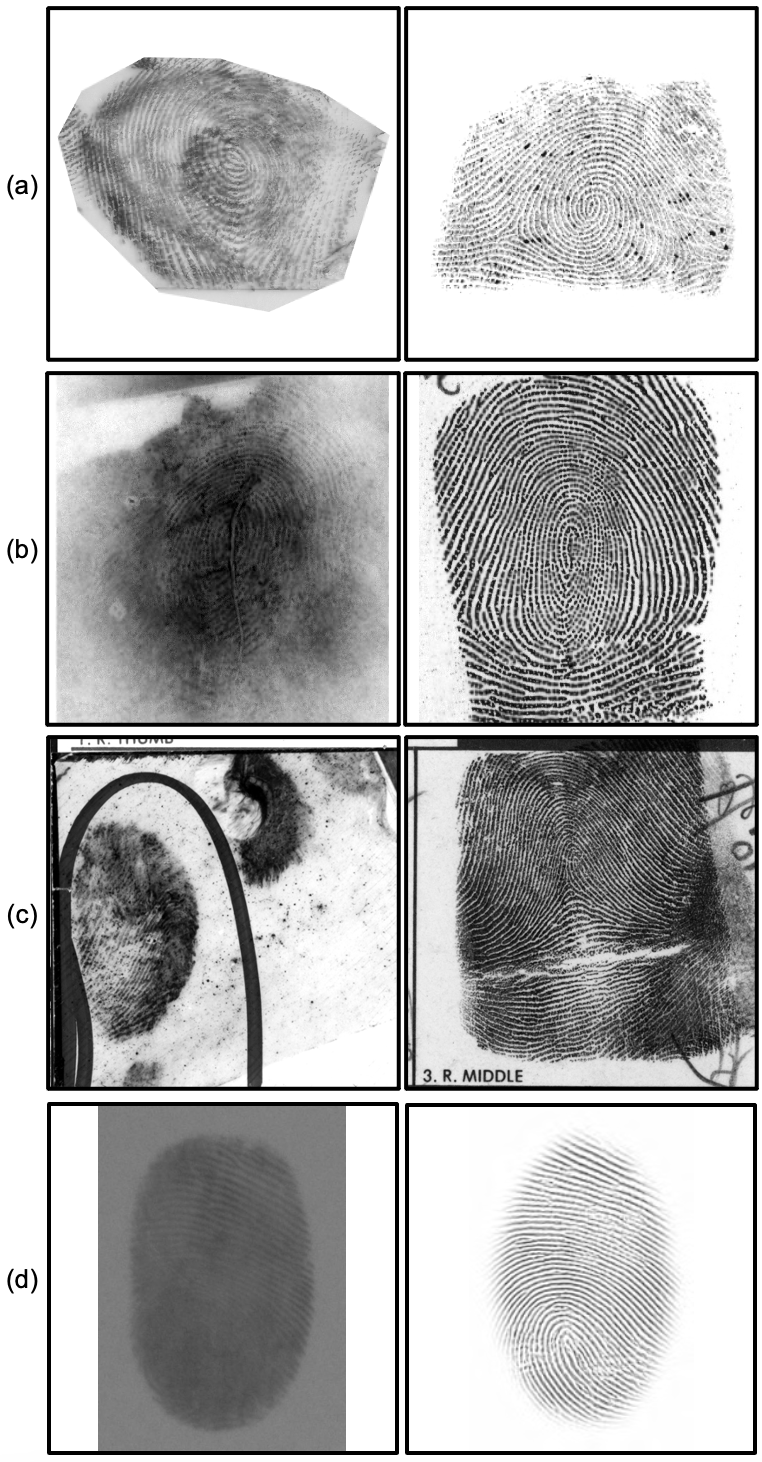} 
\caption{Example latent and corresponding rolled images from (a) N2N Latent~\cite{nist302}, (b) MSP Latent~\cite{yoon2015longitudinal}, (c) NIST SD 27~\cite{sd27}, and (d) MOLF~\cite{sankaran2015multisensor} datasets. In all the above examples, the true mate for each query latent was returned at rank-1 by the proposed method against a gallery of 100K rolled fingerprints.}
\label{fig:ex_latents}
\end{figure}

\begin{abstract}
    One of the most challenging problems in fingerprint recognition continues to be establishing the identity of a suspect associated with partial and smudgy fingerprints left at a crime scene (i.e., latent prints or fingermarks). Despite the success of fixed-length embeddings for rolled and slap fingerprint recognition, the features learned for latent fingerprint matching have mostly been limited to local minutiae-based embeddings and have not directly leveraged global representations for matching. In this paper, we combine global embeddings with local embeddings for state-of-the-art latent to rolled matching accuracy with high throughput. The combination of both local and global representations leads to improved recognition accuracy across NIST SD 27, NIST SD 302, MSP, MOLF DB1/DB4, and MOLF DB2/DB4 latent fingerprint datasets for both closed-set (84.11\%, 54.36\%, 84.35\%, 70.43\%, 62.86\% rank-1 retrieval rate, respectively) and open-set (0.50, 0.74, 0.44, 0.60, 0.68 FNIR at FPIR=0.02, respectively) identification scenarios on a gallery of 100K rolled fingerprints. Not only do we fuse the complimentary representations, we also use the local features to guide the global representations to focus on discriminatory regions in two fingerprint images to be compared. This leads to a multi-stage matching paradigm in which subsets of the retrieved candidate lists for each probe image are passed to subsequent stages for further processing, resulting in a considerable reduction in latency (requiring just 0.068 ms per latent to rolled comparison on an AMD EPYC 7543 32-Core Processor, roughly 15K comparisons per second). Finally, we show the generalizability of the fused representations for improving authentication accuracy across several rolled, plain, and contactless fingerprint datasets.
\end{abstract}

\begin{IEEEkeywords}
Latent Fingerprint Recognition, Fixed Length Embeddings, Latent Fingerprint Enhancement, Fusion of Local and Global Embeddings, Universal Fingerprint Representation
\end{IEEEkeywords}

\IEEEpeerreviewmaketitle

\section{Introduction}
\IEEEPARstart{L}{atent} fingerprints are fingerprint impressions that are left behind, unintentionally, on surfaces such as glass, metal, and plastic, and are often invisible to the human eye. However, these prints can be lifted and analyzed using specialized techniques to enhance the friction ridge patterns present in the prints~\cite{yamashita2010fingerprint}. Once the prints have been enhanced, digital images can be compared against a database of known ten-print fingerprints (rolled or slap) in law enforcement databases, which can help identify the person who left the prints behind. In fact, latent fingerprint recognition has been used as a tool in forensics and criminal investigations over the last century and are often regarded as a credible source of information in identifying suspects~\cite{handbook}; however, the reliability of automatic latent to rolled fingerprint matching considerably lags that of rolled to rolled fingerprint matching. As a result, some innocent individuals, like in the case of Brandon Mayfield~\cite{oig2006review}, have unfortunately been incarcerated due to inaccurate latent to rolled comparison by automatic fingerprint identification systems (AFIS) and failure of forensic examiners to follow the ACE-V protocol, established in the 1980s~\cite{ashbaugh1999quantitative}. As demonstrated in the four example latent fingerprints shown in Figure~\ref{fig:ex_latents}, some of the reasons for low performance in latent fingerprint recognition include poor ridge-valley contrast, occlusion, distortion, varying background, and incomplete fingerprint patterns. Because of these challenges, latent fingerprint recognition remains one of the most challenging problems in biometrics, akin to matching poor quality face images from CCTV surveillance frames to face mugshot photos.

In general, AFIS optimized for rolled or plain fingerprint impressions do not achieve comparable accuracy for latent fingerprints, even when finetuned on publicly available latent fingerprint datasets. For example, the commercial software Verifinger v12.3 achieves a true accept rate (TAR) of 99.93\% at a false accept rate (FAR) of 0.01\% on the NIST SD 14 rolled fingerprint dataset, but only a TAR of 55.04\% on NIST SD 27 latent dataset. This has sparked a number of studies that have focused on improving individual components of the fingerprint recognition pipeline to work better for latent impressions, such as those focusing on foreground segmentation of the ridge structure~\cite{cao2014segmentation}, latent enhancement~\cite{li2018deep,huang2020latent,joshi2019latent,zhu2023fingergan}, minutiae extraction~\cite{ozturk2022minnet,tang2017latent,darlow2017fingerprint,cao2018latent}, and orientation field estimation~\cite{yang2014localized,feng2012orientation}. However, few studies have focused on an end-to-end system to improve the latent to rolled fingerprint recognition pipeline, which is necessary since optimizing individual components separately may lead to sub-optimal performance when integrated together and tested as a complete system. Of those studies that do report on an end-to-end recognition system~\cite{cao2019end,cao2018automated,tang2017fingernet}, the highest rank-1 retrieval rate achieved is 65.7\%~\cite{cao2019end}, computed on 258 latent probes from NIST SD 27 against a background of 100K rolled fingerprints.


Furthermore, despite recent advancements in deep learning techniques for fixed-length representations (embeddings) for rolled/plain fingerprint recognition, these global representations have yet to be leveraged effectively for latents, likely due to the domain gap and limited availability of large-scale latent fingerprint datasets to learn such representations. Therefore, in this study, we propose an end-to-end pipeline for latent fingerprint recognition which leverages both a learned, global fingerprint representation (i.e., entire friction ridge pattern) and local representations (i.e., minutiae and virtual minutiae\footnote{Virtual minutiae are densely sampled points on an evenly spaced grid on the extracted fingerprint ridge area.}) for improved accuracy and search speed of latent to rolled fingerprint recognition. We not only fuse these complimentary local and global embeddings, but utilize the local features to inform or guide the global representations to focus on discriminatory regions of the input fingerprint image pairs for an improved matching accuracy.

Lastly, existing latent to rolled fingerprint matching pipelines are highly tuned specifically for latent fingerprints, whereas our representation and matching pipeline is generalizable and effective across a wide range of fingerprint sensors (e.g., optical, capacitive, etc.) and image domains (e.g., latent, rolled, plain, contactless captures via mobile phone cameras, etc.). In a sense, we report on a \textit{universal fingerprint representation} that is agnostic to fingerprint sensors and fingerprint capture mode. Concretely, the contributions of this research are the following:

\begin{enumerate}
    \item Design of an end-to-end latent fingerprint recognition pipeline using deep learning methods, including algorithms for segmentation, enhancement, minutiae extraction, and a fusion of global and local embeddings.
    \item State-of-the-art (SOTA) latent to rolled/plain fingerprint search across multiple datasets, including NIST SD 27~\cite{sd27}, NIST SD 302 Latents (N2N Latents)~\cite{nist302}, MSP Latent~\cite{yoon2015longitudinal}, and MOLF datasets~\cite{sankaran2015multisensor}.
    \item Faster search speed (low latency) due to our multi-stage search scheme, while maintaining SOTA recognition accuracy for both closed-set and open-set identification. 
    \item Generalization of representation (embedding) from LFR-Net is shown via SOTA authentication performance across several rolled (NIST SD 14~\cite{sd14}), plain (NIST SD 302~\cite{nist302}), and contact to contactless fingerprint matching datasets (PolyU Contactless 2D to Contact-based 2D~\cite{lin2018matching} and ZJU Finger Photo and Touch-based~\cite{grosz2021c2cl}) using the same network, a step toward a universal fingerprint recognition system.
\end{enumerate}

\begin{table*}[t]
\renewcommand{\arraystretch}{1.3}
\caption{Summary of recently published latent fingerprint recognition studies.}
\label{tab:prior_work}
\begin{tabular}{c|l|l|c}
\noalign{\hrule height 1.5pt}
\textbf{Study} & \specialcell{\textbf{Approach}} & \specialcell{\textbf{Database}} & \specialcell{\textbf{Rank-1}}\\
\noalign{\hrule height 1.0pt}
\specialcell{FingerNet, 2017~\cite{tang2017fingernet}} & \leftcell{CNN methods for minutiae extraction, orientation field estimation, segmentation,\\ and enhancement. Search results on a gallery of 40K.} & \specialcell{NIST SD 27~\cite{sd27}} & \specialcell{$\sim$35.0\%}\\
\hline
\specialcell{MSU-AFIS, 2019~\cite{cao2019end}} & \leftcell{CNN methods for enhancement, segmentation, minutiae, and virtual minutiae\\extraction. Search results on a gallery of 100K.} & \specialcell{NIST SD 27~\cite{sd27}\\MSP Latent$^\dagger$~\cite{yoon2015longitudinal}} & \specialcell{65.7\%\\69.4\%}\\
\hline
\specialcell{Gu et al., 2020~\cite{gu2020latent}} & \leftcell{Registration of latents via matching dense, undirected sampling points + virtual\\minutiae (not a fully automated system). Search results on a gallery of 100K.} & \specialcell{NIST SD 27~\cite{sd27}\\ MOLF DB3/DB4~\cite{sankaran2015multisensor}} & \specialcell{70.1\%\\19.8\%}\\
\hline
\specialcell{MinNet, 2022~\cite{ozturk2022minnet}} & \leftcell{CNN-based minutiae patch embedding network + local similarity assignment (LSA)\\algorithm for matching. Search results on a gallery of 5,560, 316, and 168 images\\from EGM, FVC-Latent, and Tshingua-Latent databases, respectively.} & \specialcell{EGM (private dataset)\\FVC-Latent~\cite{ozturk2022minnet}\\Tshingua-Latent~\cite{ozturk2022minnet}} & \specialcell{92.39\%\\95.57\%\\99.40\%}\\
\hline
\specialcell{FingerGAN, 2023~\cite{zhu2023fingergan}} & \leftcell{GAN-based enhancement + Verifinger v12.1. Search results on a gallery of 27,258.} & \specialcell{NIST SD 27~\cite{sd27}\\MOLF DB1/DB4~\cite{sankaran2015multisensor}\\MOLF DB2/DB4~\cite{sankaran2015multisensor}\\MOLF DB3/DB4~\cite{sankaran2015multisensor}} & \specialcell{59.69\%\\25.34\%\\22.23\%\\29.43\%}\\
\noalign{\hrule height 1.0pt}
\specialcell{\textbf{LFR-Net}\\\textbf{(proposed approach)}} & \leftcell{CNN-based latent enhancement, segmentation, and fusion of local (minutiae + virtual\\minutiae) and global (AFR-Net~\cite{grosz2022afr}) embeddings for matching. Search results\\reported on a gallery of 100K.} & \specialcell{NIST SD 27~\cite{sd27}\\N2N Latent~\cite{nist302}\\MSP Latent~\cite{yoon2015longitudinal}\\ MOLF DB1/DB4~\cite{sankaran2015multisensor}\\MOLF DB2/DB4~\cite{sankaran2015multisensor}} & \specialcell{84.11\%\\54.36\%\\84.35\%\\70.43\%\\62.86\%}\\
\noalign{\hrule height 1.5pt}
\multicolumn{4}{p{0.95\linewidth}}{$^\dagger$ Did not specify the test split.}\\
\end{tabular}
\end{table*}

\section{Related Work}
\subsection{Latent Fingerprint Enhancement}
A critical step in improving the accuracy of latent to rolled comparison is alleviating the effect of various degradations present in latent fingerprints through preprocessing aimed at enhancing the contrast of the latent fingerprint ridge structure. A multitude of latent enhancement methods have been proposed over the years, ranging from classical computer vision techniques~\cite{cappelli2009semi, chikkerur2007fingerprint, yoon2011latent, feng2012orientation, yang2014localized, zhang2012latent, zhang2013adaptive} to state of the art deep learning methods~\cite{tang2017fingernet, cao2015latent, svoboda2017generative, li2018deep, liu2020automatic, dabouei2018id, joshi2019latent, huang2020latent, zhu2023fingergan}. 

Early enhancement efforts utilized contextual filtering and directional filtering~\cite{cappelli2009semi,chikkerur2007fingerprint}, but these methods were limited in their effectiveness for enhancing latent fingerprints due to corrupted ridge structures and unreliable orientation and frequency estimation compared to that of plain and rolled fingerprints. This led to many subsequent studies on improving the ridge orientation estimation for latent fingerprints. For example, Yoon et al.~\cite{yoon2011latent} utilized a combination of polynomial models and Gabor filters to improve latent orientation estimation. Similarly, Feng et al.~\cite{feng2012orientation} utilized an orientation patch dictionary and Gabor filters for latent enhancement and Yang et al.~\cite{yang2014localized} extended this approach by utilizing local orientation dictionaries, which increased the flexibility of the approach to find better orientation fields. However, the variance in ridge frequency of distorted latent fingerprints limited the utility of these methods in improving overall matching accuracy. 

Subsequent efforts introduced deep neural networks to improve the enhancement of latent fingerprints. In addition to a combination of short-time Fourier transform (STFT) and Gabor filters, Cao et al.~\cite{cao2015latent} trained a convolutional neural network (CNN) autoencoder to enhance latent fingerprints. Variants of the CNN-based approach were also proposed in~\cite{svoboda2017generative, li2018deep, liu2020automatic}. Generative adversarial networks (GANs) have also been adopted for latent fingerprint enhancement and these methods have shown promise in restoring ridge and valley structures~\cite{dabouei2018id, joshi2019latent, huang2020latent, zhu2023fingergan}. However, as shown in Table~\ref{tab:minu_stats}, these methods have a tendency to hallucinate ridge lines and produce spurious minutiae that may degrade matching performance. Furthermore, critical to the success of many of these methods was access to large databases of mated rolled and latent fingerprint image pairs for training, many of which are unfortunately not publicly available to other researchers. In this work, we adopt the efficient CNN architecture of Squeeze U-Net~\cite{beheshti2020squeeze} for latent enhancement without access to any latent training data. Instead, we employ a series of data augmentations on a dataset of rolled and plain fingerprint impressions in order to mimic the degradations present in latent fingerprints and our network is trained to restore the degraded images to their original input. A comparison between the performance of our enhancement network and several previous baselines is given in section~\ref{sec:ablation_study}.

\subsection{Latent Fingerprint Recognition}
Despite recent success of deep learning global representations for fingerprint matching, all existing latent fingerprint recognition systems (to the best of our knowledge) utilize minutiae-based matchers for computing final similarity scores between latent and rolled image pairs. For example, Cao et al.~\cite{cao2019end} and \"Ozt\"urk et al.~\cite{ozturk2022minnet} utilize variants of the local similarity assignment algorithm proposed in \cite{cappelli2010minutia} for computing minutiae similarity scores, Tang et al.~\cite{tang2017fingernet} utilized the extended clique model for minutiae matching, FingerGAN used Verifinger v12.1 for matching and Gu et al.~\cite{gu2022latent} utilized multi-scale fixed-length embeddings for indexing to reduce the potential candidate list in combination with MSU-AFIS~\cite{cao2019end} for computing the similarity scores. Even though deep learning networks are used within many of these minutiae-based methods to produce local minutiae descriptors around minutiae points, no existing study directly leveraged a global embedding as an additional similarity comparison. In this paper, we propose to use a global embedding score for improving the latent to rolled matching performance, in conjunction with local minutiae embeddings for minutiae matching. A table giving a brief summary of the recent publications on latent fingerprint recognition is given in Table~\ref{tab:prior_work}

\begin{figure*}[!t]
\centering
\includegraphics[width=0.95\linewidth]{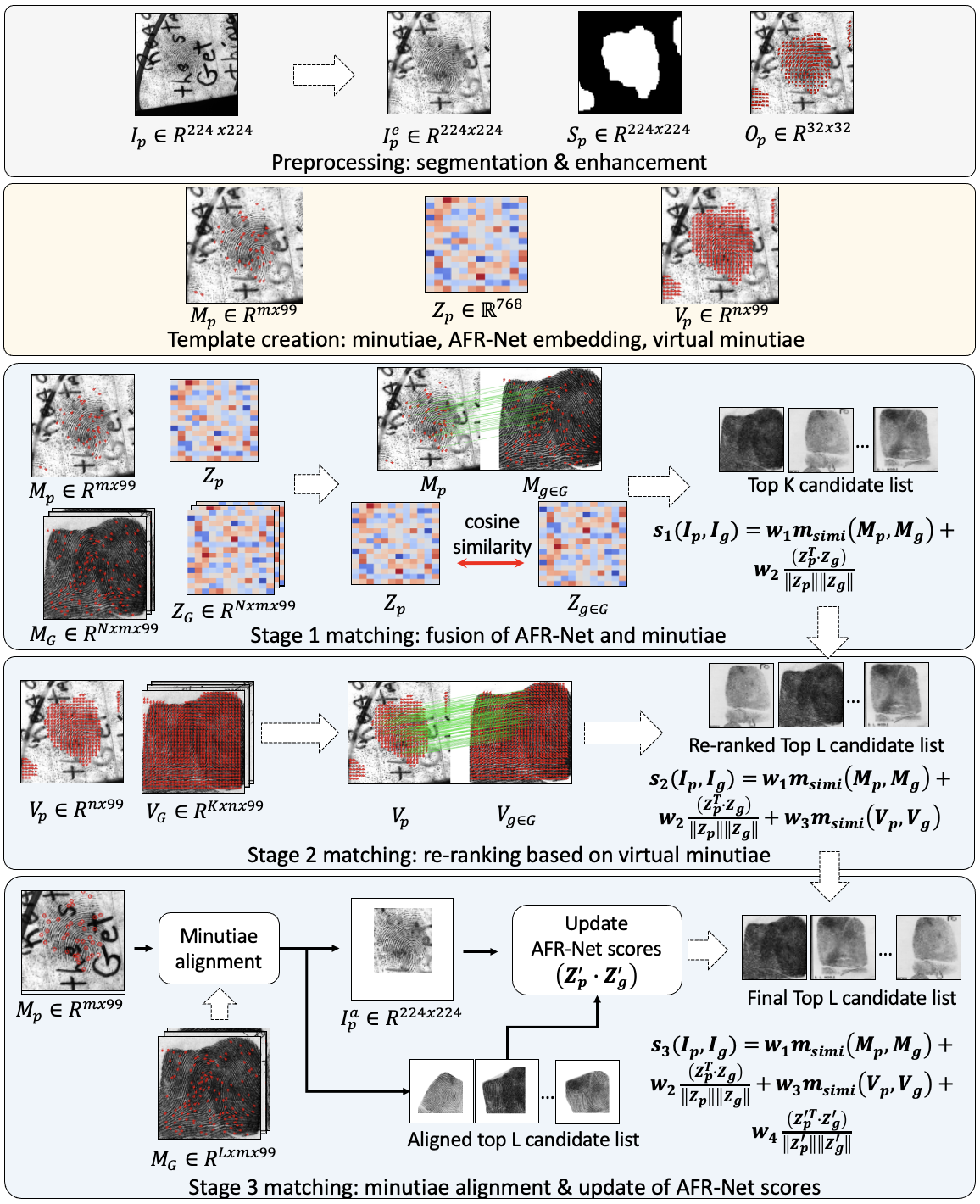} 
\caption{Overview of LFR-Net. An input latent probe image $I_p$ is first automatically segmented and enhanced to generate $I^{e}_p$, orientation field $O_p$, and segmentation mask $S_p$. Then, $I^{e}_p$ is passed to a minutiae extraction network, minutiae descriptor network, and AFR-Net~\cite{grosz2022afr} to produce a minutiae feature set $M_p$, virtual minutiae feature set $V_p$, and AFR-Net embeddings $Z_p$, which are embedded into a template for matching $(M_p,Z_p,V_p)$. Once extracted, the probe feature template is compared with each gallery template $(M_g,Z_g,V_g)$ in the gallery $G$ of size $N$ via a similarity function $s(I_p, I_g)$ in three stages. The final output after the third matching stage is a candidate list of L candidates.}
\label{fig:overview}
\end{figure*}

\section{LFR-Net: Latent Fingerprint Recognition Network}
Our approach for accurate and efficient latent fingerprint search consists of a combination of local (minutiae and virtual minutiae) and global features (AFR-Net embeddings~\cite{grosz2022afr}). Additionally, due to the low contrast, occlusion, and varying background present in many latent fingerprint images, we first incorporate automatic segmentation and enhancement of  latent fingerprint images prior to feature extraction. The following sections will describe each component of our latent to rolled fingerprint matcher, referred to as LFR-Net. These components include enhancement, segmentation, minutiae extraction, virtual minutiae extraction, global embedding, realignment for improved global embeddings, and a multi-stage search strategy. An overview of the pipeline is illustrated in Figure~\ref{fig:overview}.


\begin{figure*}
\includegraphics[width=\linewidth]{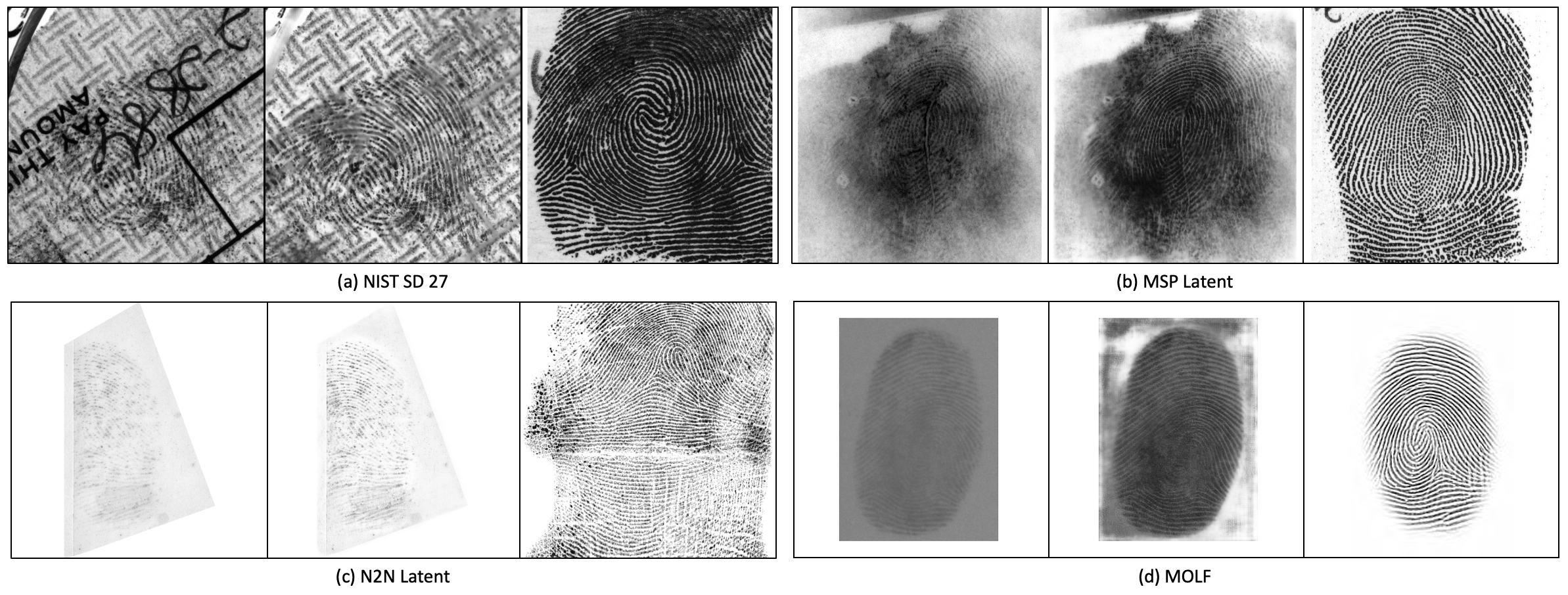} 
\caption{Example enhanced latent images from (a) NIST SD 27~\cite{sd27}, (b) MSP Latent dataset~\cite{yoon2015longitudinal}, (c) N2N Latent~\cite{nist302}, and (d) MOLF~\cite{sankaran2015multisensor} datasets. In each subfigure, the left image is the original latent image, the middle image is the enhanced latent image using the proposed enhancement network, and the right image is the corresponding rolled mate.}
\label{fig:ex_enhancements}
\vspace{-1.5em}
\end{figure*}

\subsection{Latent Enhancement and Segmentation}
The terminology introduced for NIST SD 27 denotes the quality of latent fingerprints as either good, bad, or ugly depending on several factors, including the percentage of the fingerprint ridge structure occluded, noise obscuring the ridge structure, and low contrast of the ridges compared to the background content of the image. To make things even more challenging, the quality and appearance of latent fingerprints can vary drastically across different databases, either collected in the lab (as is the case for the NIST SD 302 (N2N)~\cite{nist302} and IIIT-D MOLF datasets~\cite{sankaran2015multisensor}) or from real crime scenes (as is the case for NIST SD 27~\cite{sd27} and MSP Latent datasets~\cite{yoon2015longitudinal}). Therefore, latent enhancement is a critical yet challenging step for accurate and reliable latent to rolled fingerprint matching. See Figure~\ref{fig:ex_enhancements} for example latent and rolled/plain fingerprint pairs showcasing the various differences between latent datasets.

\begin{figure}
\includegraphics[width=\linewidth]{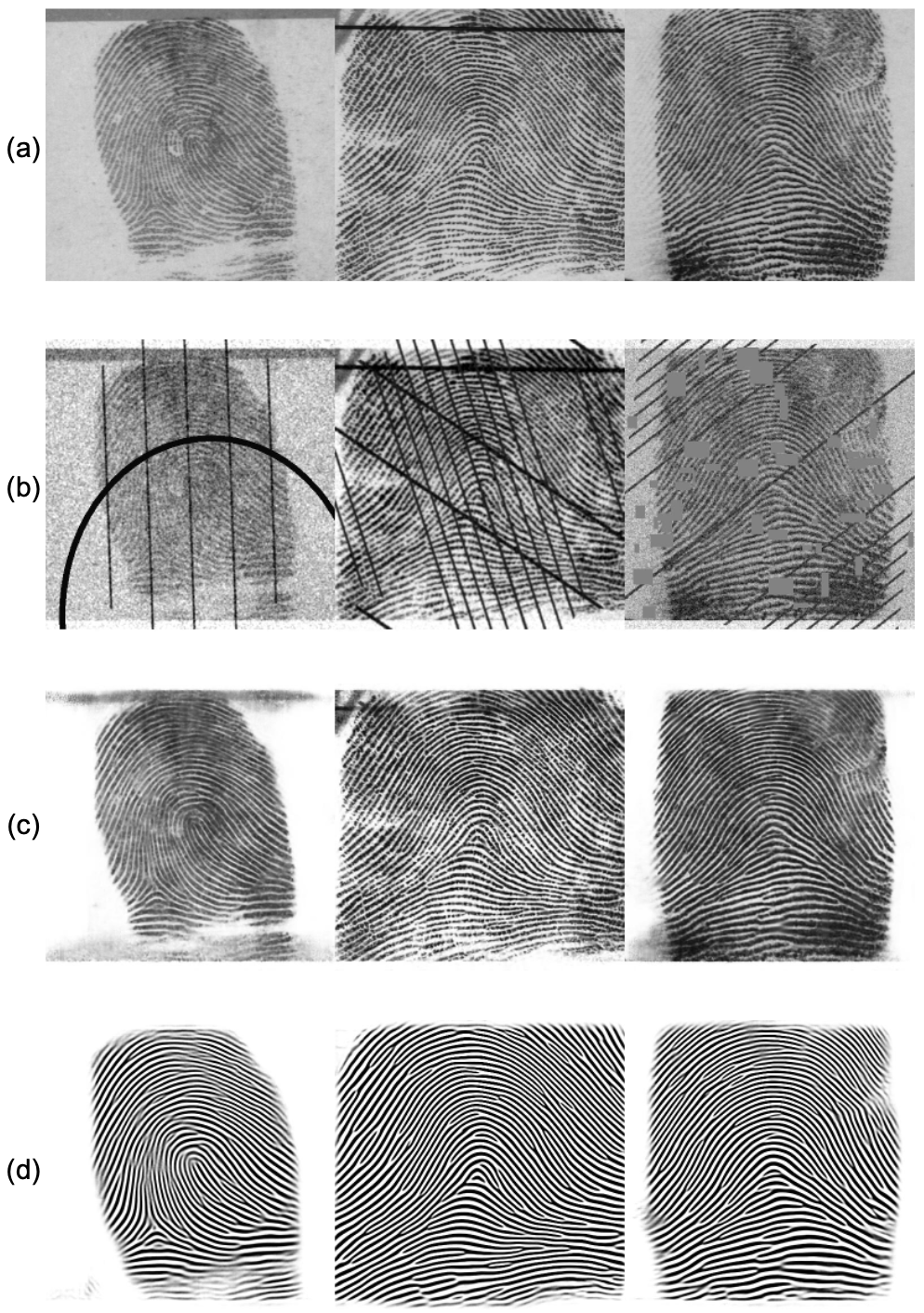} 
\caption{Example data augmentations to train the latent enhancement network. Random Gaussian blurring, Gaussian noise, downsampling, partial occlusions, and contrast adjustments are applied to rolled fingerprint images to generate low quality fingerprints that resemble characteristics of latent fingerprints. (a) original rolled fingerprints, (b) simulated latent fingerprints after data augmentations, (c) predicted enhanced output, and (d) predicted binary ridge image.}
\label{fig:data_augs}
\end{figure}

To address the problem of latent enhancement, we focus on two key factors degrading the quality of latent prints; namely, presence of noise occluding areas of the latent fingerprint ridge structure and low contrast of the ridges. To remove noise from the latent images, we train a de-noising CNN network to remove noise and fill-in occluded regions of the fingerprint ridge structure. This network architecture is modeled from Squeeze U-Net~\cite{beheshti2020squeeze}, an efficient network proposed for image segmentation, where we have adapted it for latent enhancement. Next, we aim to highlight the ridge structure of the latent fingerprints by constraining the network to segment the fingerprint ridge lines from the background. To accomplish this, we introduce an additional channel to the output of our enhancement network and optimize for both tasks in a single architecture. Thus, the output of the enhancement network is two channels, one for the enhanced image and another for the ridge lines. Note, the outputs of both channels are gray-scale and in the range [0,255]. A few example enhancement outputs from this network are shown in the middle column of each sub-figure in Figure~\ref{fig:ex_enhancements} and the bottom two rows of Figure~\ref{fig:data_augs}.

\begin{figure}
\includegraphics[width=\linewidth]{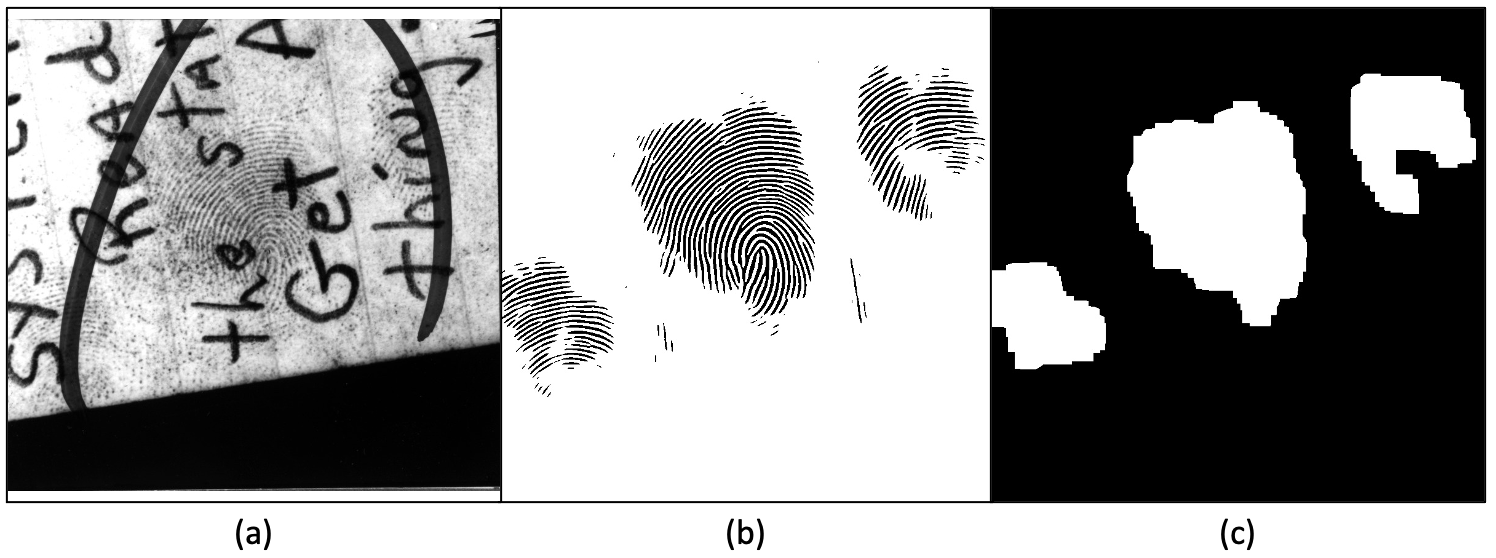} 
\caption{Example mask prediction for a latent image from NIST SD 27. (a) input latent image, (b) gray-scale ridge image output by the enhancement network, and (c) binary mask obtained after a series of Guassian blurring, thresholding, and morphological operations on (b).}
\label{fig:segmentation}
\end{figure}

To locate and segment the latent fingerprint area from the background image content, we use the predicted fingerprint ridges as a segmentation mask for localizing the latent fingerprint area by performing a series of simple image processing operations. First, a Gaussian filter with kernel size (5,5) is applied to the predicted ridge map, followed by a thresholding operation with a threshold of 150 on the pixel values to obtain the binary ridge lines in the range [0,1]. Next, a morphological closing operation with a kernel size of (9,9) is repeated 3 times, followed by three morphological opening operations with a kernel size of (9,9). Finally, the mitigate erroneous predictions, the resulting mask defaults to the entire image if the resulting mask after processing has an area of less than 10,000 pixels. Since our enhancement network is fully convolutional, it can accept images of any resolution; however, the final segmented images are cropped to a height and width of $512\times512$ pixels at a resolution of 500 ppi. Figure~\ref{fig:segmentation} illustrates the process of converting a predicted gray-scale ridge image to a binary segmentation mask for an example latent fingerprint from NIST SD 27.

Due to a lack of publicly available large-scale latent databases, we utilize several data augmentations to mimic the distribution of latent fingerprints using a collection of rolled and slap fingerprints. These data augmentations are illustrated in (b) of Figure~\ref{fig:data_augs} and consist of random amounts of Gaussian blurring, Gaussian noise, downsampling, partial occlusions, and contrast adjustments. The enhancement network is trained to remove these degradations from the augmented images via an MSE loss between the predicted, enhanced image and the original, unperturbed image. Furthermore, we compute an additional MSE loss between the predicted ridge images and the ridge images extracted from the original input fingerprints via Verifinger v12.3 (normalized to the range [0,255]). Equal weight is given to the two MSE loss terms during training.

\begin{figure*}
\includegraphics[width=\linewidth]{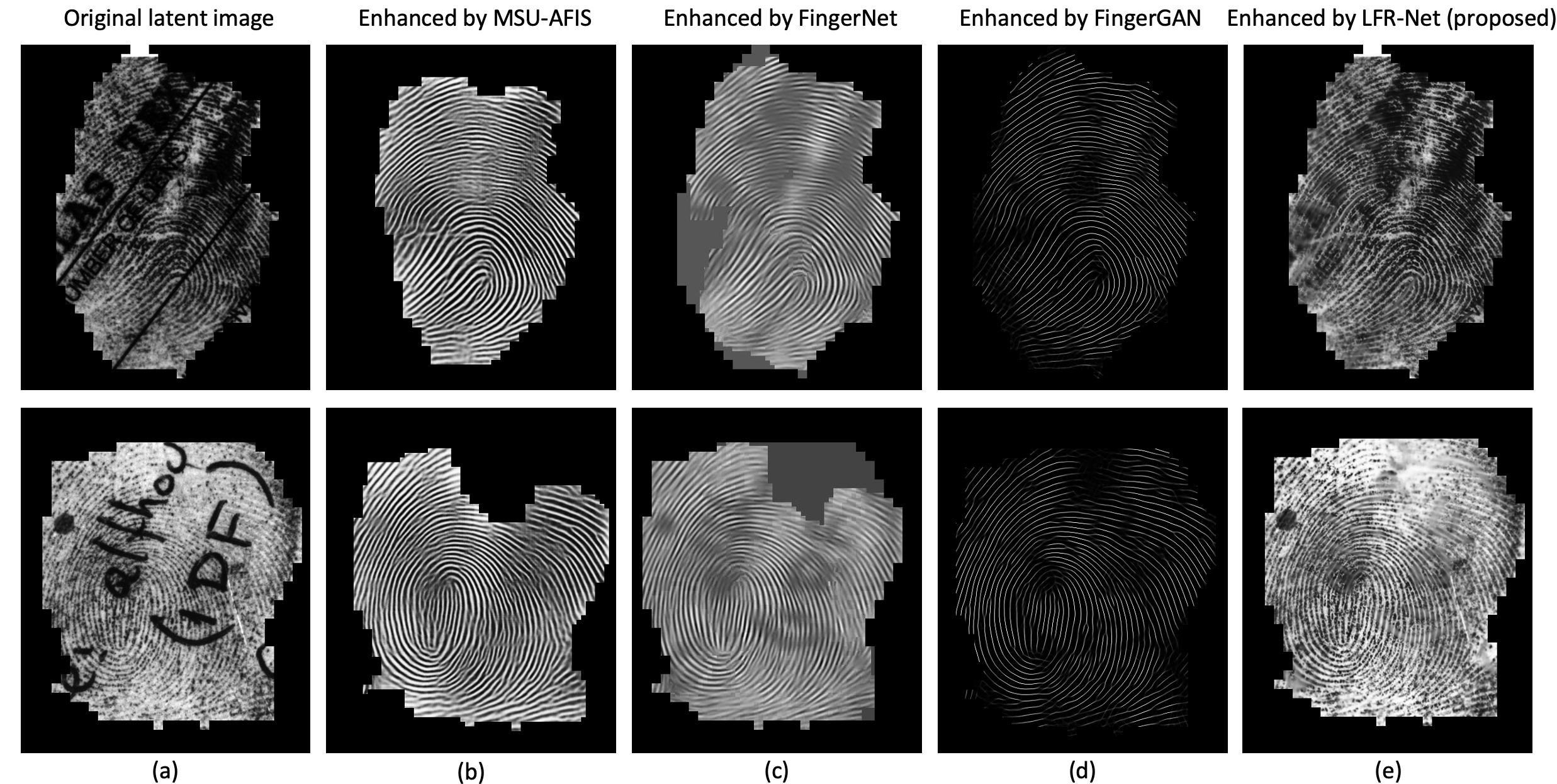} 
\caption{Two example comparisons of several baseline enhancement algorithms and the proposed enhancement network. (a) original latent images, (b) enhanced by MSU-AFIS~\cite{cao2019end}, (c) enhanced by FingerNet~\cite{tang2017fingernet}, (d) enhanced by FingerGAN~\cite{zhu2023fingergan}, and (e) enhanced by LFR-Net (proposed).}
\label{fig:enhancement_comparison}
\end{figure*}

The enhancement network is trained on the MSP longitudinal fingerprint dataset (rolled fingerprints only)~\cite{yoon2015longitudinal}, a subset of NIST SD 302 (rolled and plain fingerprints only)~\cite{nist302}, and a dataset of plain fingerprint impressions referred to as the MSU Self-Collection. Details on number of fingers/images contained in each of these datasets are provided in Table~\ref{tab:datasets}. Ground truth binary images for all the training images are obtained using Verifinger v12.3. The network was trained on 2 Nvidia RTX A6000 GPUs for 11 epochs utilizing an initial learning rate of 0.001, polynomial learning rate schedule, and Adam optimizer. As is shown in section~\ref{sec:ablation_study}, despite not being trained on any real latent images, our enhancement network is able to outperform many of the existing latent enhancement methods in the literature. For illustration, example enhancements from each of these methods is shown in Figure~\ref{fig:enhancement_comparison}.

\subsection{Minutiae Extraction}
Our minutiae extraction network consists of a ResNet50 backbone, self-attention transformer layers, and a series of transpose convolutional layers to predict a 12-channel minutiae map, a representation for minutiae points introduced in \cite{cao2019end}. This minutiae map is converted to a list of $(x,y,\theta)$ locations for each minutiae point and a set of $96\times96$ image patches centered around each minutiae are aligned based on the orientation $\theta$ and fed into a separate ResNet50 model to extract a set of descriptors associated with each minutiae. These descriptors are each 96-dimensional and used in the minutiae similarity calculation when comparing two sets of minutiae points extracted from a given fingerprint image pair. Thus, in conjunction with the $(x,y,\theta)$ locations of each minutiae point and assuming $m$ minutiae points in total, a given minutiae template $M$ will be of dimension $M\in \mathcal{R}^{m\times99}$. The architecture details of the minutiae extraction network are given in Table~\ref{tab:minu_architecture}.

\begin{table}[h]
\caption{Architecture details for the minutiae extractor network. Batch normalization and ReLU activation are applied after each convolution, except for the last layer which uses a Sigmoid activation.}
\label{tab:minu_architecture}
\begin{tabularx}{\linewidth}{X || M || X}
\noalign{\hrule height 1.5pt}
\textbf{Layer Type} & \textbf{Output Dim.} & \textbf{Parameters} \\
\noalign{\hrule height 1.0pt}
Conv2d & 64\times112\times112 & k=7x7, padding=3, stride=2\\
\hline
Conv2d & 256\times56\times56 &
    $\begin{bmatrix}
        \text{k=}1\times1, \text{ch=}64 \\
        \text{k=}3\times3, \text{ch=}64 \\
        \text{k=}1\times1, \text{ch=}256
    \end{bmatrix}$x3\\
\hline
Conv2d & 512\times28\times28 &
    $\begin{bmatrix}
        \text{k=}1\times1, \text{ch=}128 \\
        \text{k=}3\times3, \text{ch=}128 \\
        \text{k=}1\times1, \text{ch=}512
    \end{bmatrix}$x4\\
\hline
Conv2d & 1024\times14\times14 &
    $\begin{bmatrix}
        \text{k=}1\times1, \text{ch=}256 \\
        \text{k=}3\times3, \text{ch=}256 \\
        \text{k=}1\times1, \text{ch=}1024
    \end{bmatrix}$x6\\
\hline
MLP & 384\times196 & in=1024, hid=1024, out=384 \\
\hline
\specialcell{Self-Attention + MLP} & 384\times196 & in=384, hid=1536, out=384 \\
\hline
\specialcell{Conv2d Transpose} & 384\times28\times28 &
    k=2x2, stride=2 \\
    \hline
Conv2d & 384\times28\times28 &
    $\begin{bmatrix}
        \text{k=}3\times3, \text{ch=}384 \\
    \end{bmatrix}$x2\\
\hline
\specialcell{Conv2d Transpose} & 192\times56\times56 &
    k=2x2, stride=2 \\
\hline
Conv2d & 192\times56\times56 &
    $\begin{bmatrix}
        \text{k=}3\times3, \text{ch=}192 \\
    \end{bmatrix}$x2\\
\hline
\specialcell{Conv2d Transpose} & 96\times112\times112 &
    k=2x2, stride=2 \\
\hline
Conv2d & 96\times112\times112 &
    $\begin{bmatrix}
        \text{k=}3\times3, \text{ch=}96 \\
    \end{bmatrix}$x2\\
\hline
\specialcell{Conv2d Transpose} & 48\times224\times224 &
    k=2x2, stride=2 \\
\hline
Conv2d & 48\times224\times224 &
    $\begin{bmatrix}
        \text{k=}3\times3, \text{ch=}48 \\
    \end{bmatrix}$x2\\
\hline
Conv2d & 12\times224\times224 & k=1x1\\
\noalign{\hrule height 1.5pt}
\end{tabularx}
\end{table}

For matching minutiae points, we compute a similarity matrix between all Euclidean normalized minutiae descriptors and utilize the local similarity with relaxation (LSS-R) algorithm (as described in Minutiae Cylinder-Code (MCC) \cite{cappelli2010minutia}) to refine and remove false correspondences. Finally, the cosine similarity between the descriptors of corresponding minutiae points are summed to yield a final minutiae similarity score. Due to the nature of latent fingerprint formation, we find it is extremely useful to align the minutiae points prior to extracting the minutiae descriptors. This step imparts the similarity calculation with rotation invariance, a critical factor in unconstrained latent fingerprint recognition.

\begin{figure}
\includegraphics[width=\linewidth]{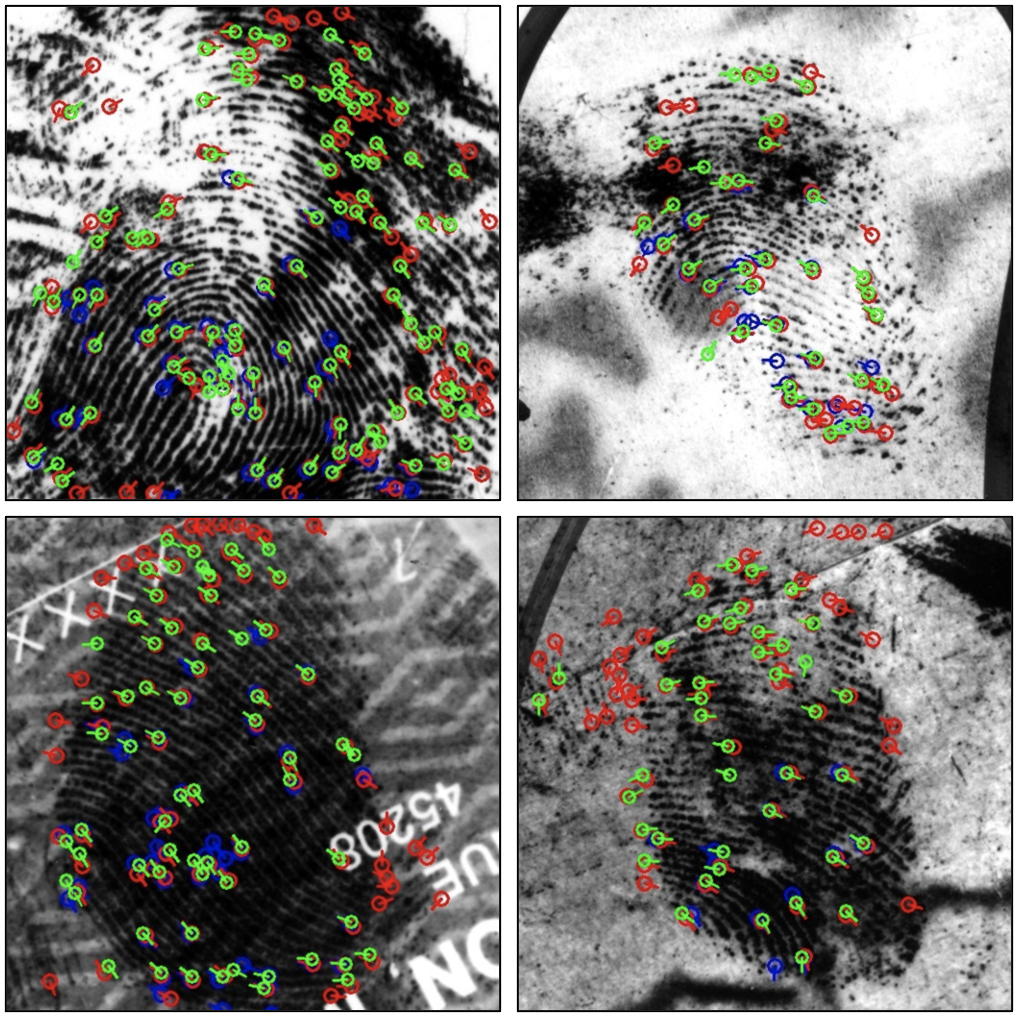} 
\caption{Visual comparison of minutiae extracted by our method (shown in green), Verifinger v12.3 (shown in red) and manually marked minutiae (shown in blue). Best viewed in color.}
\label{fig:minu_comparison}
\end{figure}

Our minutiae extraction and descriptor networks are trained on the MSP (rolled fingerprints only), NIST SD 302 (rolled and plain fingerprints only), and MSU Self-Collection (plain fingerprints only) training datasets. An MSE loss between predicted and ground truth minutiae points (obtained using the commercial Innovatrics v2.4.10 SDK) was used to supervise the minutiae extraction network. For training the minutiae descriptor model, minutiae patches of size 96$\times$96 pixels were extracted from corresponding minutiae points between multiple impressions of each finger in the training set. To ensure reliability of ground truth corresponding minutiae patches, only corresponding minutiae points common among all impressions of the same finger were used and assigned a label for training. The Additive Angular Margin (ArcFace) loss function was used to supervise the descriptor model in classifying image patches belonging to the same minutiae point~\cite{deng2019arcface}. Both networks were trained on 4 Nvidia RTX A6000 GPUs for 56 epochs, with an initial learning rate of 0.0001, polynomial learning rate schedule, and Adam optimizer. A visual comparison of four example latent images annotated with minutiae from our minutiae extractor (shown in green), Verifinger v12.3 (shown in red), and manually marked minutiae (shown in blue) is provided in Figure~\ref{fig:minu_comparison}. Due to the difficulty in manually marking latent minutiae points, usually very few minutiae are manually annotated. On the other hand, automatic minutiae extractors tend to detect many false (e.g., spurious) minutiae due to noise in the image. Nonetheless, compared to Verifinger, our method is detecting less spurious minutiae (as can be seen in the bottom two examples of Figure~\ref{fig:minu_comparison}).

\subsection{Virtual Minutiae Extraction}
Due to the severely low quality of the ridges in many latent fingerprints, minutiae extraction is often unreliable and may produce many spurious minutiae and/or fail to extract any minutiae points at all. Therefore, in order to enforce local features within the image as part of matching, we utilize virtual minutiae, originally suggested in \cite{cao2019end}. These virtual minutiae points are evenly spaced throughout the fingerprint area and use the estimated orientation field within the neighborhood of each point as the orientation assigned to each virtual minutiae point. Through an ablation study presented in section~\ref{sec:ablation_study}, we show the importance/utility of incorporating virtual minutiae into our pipeline.


For extracting virtual minutiae, we place a grid of virtual minutiae points at each (x,y) location of the segmented fingerprint area, separated by 16 pixels (in both x and y directions). The orientation of each 16x16 patch assigned to each virtual minutiae is estimated using the orientation field extraction algorithm described in~\cite{chikkerur2007fingerprint}. Aligned image patches centered around each virtual minutiae are then fed to the same minutiae descriptor model described above to extract embeddings for each virtual minutiae. Since we are using the same minutiae descriptor extraction network, no additional training is required to obtain the virtual minutiae points. Assuming $n$ virtual minutiae points are extracted in total, a given virtual minutiae template $V$ will be of dimension $V\in \mathcal{R}^{n\times99}$. The virtual minutiae similarity calculation between two virtual minutiae templates also utilizes the LSS-R matching algorithm~\cite{cappelli2010minutia}.

\subsection{Global Embedding Representation}
For our global representation, we utilize the recently proposed AFR-Net architecture~\cite{grosz2022afr}, which achieved high performance across a wide range of fingerprint types (rolled, plain, contactless) and sensors (optical, capacitive, etc.). AFR-Net is a combination of both CNN and ViT image recognition architectures, consisting of a shared CNN backbone and two separate classification heads (one CNN-based and the other utilizing attention blocks from ViT). The output of AFR-Net is two embeddings ($Z_a$ and $Z_c$) of 384-dimensions each and the similarity score calculation is performed via a weighted sum of the normalized dot product between both embeddings of a fingerprint pair. For simplicity, we denote the AFR-Net embeddings as $Z$, a concatenation of the two individual embeddings (764-dimensional).

AFR-Net is trained on a diverse training set consisting of a combination of rolled fingerprints~\cite{yoon2015longitudinal, sd4,engelsma2022printsgan}, plain (i.e., slap) fingerprints~\cite{grosz2022spoofgan}, mixture of rolled and plain fingerprints~\cite{nist302}, contactless (e.g., from mobile phone cameras) fingerprints~\cite{ericson2015evaluation, deb2018matching, birajadar2019towards}, and synthetic latent fingerprints~\cite{wyzykowski2022synthetic}. In total there are about 1.3 million images from 96,556 unique finger identities in training. Due to the lack of publicly available latent fingerprint datasets, we do not train on any real latent fingerprint databases and reserve the few latent dataset that we do have for evaluation. Interested readers are referred to \cite{grosz2022afr} for complete training dataset details.

\subsection{Minutiae Alignment of Global Embeddings}
As proposed in \cite{grosz2022afr}, a strategy for improving the fingerprint representations obtained via deep learning networks is to align the regions of interest between two input images, remove background and other non-overlapping regions of the fingerprint areas in both images, and pass the aligned images back into the embedding network to yield new ``refined" representations. In contrast to \cite{grosz2022afr}, where the local embeddings used to find corresponding regions of interest in both images are from an intermediate layer in the AFR-Net architecture, we directly use the minutiae correspondence between two images to compute the affine transformation which best aligns the image pair. In a sense, we are informing the global representation to focus on regions of the images which share many local similarities, in order to better distinguish between genuine pairs and close imposters.


\subsection{Multi-Stage Search Strategy}
Each of the feature sets in LFR-Net adds complimentary information for improving the reliability of a potential match, yet incurs an additional latency cost per match, which can be prohibitively expensive on a large gallery size (e.g., N=100,000). Typically, computing the similarity between global, fixed length feature vectors (such as AFR-Net embeddings), is extremely fast compared to local feature matching (e.g., minutiae graph similarity computation); however, performance on small area latent fingerprints suffers without the use of local features. Therefore, we propose a multi-stage search paradigm which reduces the size of the returned candidate list before invoking expensive local feature matching (e.g., virtual minutiae similarity computation) to refine the final ranked candidate list. 

Specifically, our hierarchical matching procedure consists of three stages. First, we return the top K (e.g., K=1,000) candidate matches using a fusion of AFR-Net similarity and minutiae matching. Next, we re-rank the top K candidates using virtual minutiae matching and obtain a smaller candidate list of size L (e.g., L=500). Finally, we align each probe image to each of its L candidate gallery images (using an affine transformation computed from corresponding minutiae points) and obtain a new set of AFR-Net embeddings on the aligned images in order to further refine the final candidate list. An illustration of this multi-stage search strategy is shown in Figure~\ref{fig:overview}. A discussion on the latency savings utilizing our three stage match procedure is given in section~\ref{sec:latency}. The scores after each stage of matching are normalized to the range [0,1] based on a set of weights ($w_1$=0.4, $w_2$=0.4, $w_3$=0.18, and $w_4$=0.02) determined empirically on a validation set of latent fingerprints from the MSP latent database (which is separate from the MSP latent test dataset). The overall algorithm for LFR-Net is given in Algorithm~\ref{alg:lfrnet}.

\begin{algorithm}
\caption{Return a ranked candidate list, given an input latent fingerprint probe ($I_p$) and gallery of rolled fingerprint images ($I_G$) using the proposed LFR-Net matcher.}\label{alg:lfrnet}

\begin{algorithmic}[1]
\Procedure{Match}{$I_p,I_G$}

\State \textit{//  Initialize score weights}
\State $w_1, w_2, w_3, w_4 := 0.4, 0.4, 0.18, 0.02$
\State

\State \textit{//  Initialize no. candidates passed to $2^{nd}$ stage.}
\State $K := 1000$
\State

\State \textit{//  Initialize no. candidates passed to $3^{rd}$ stage.}
\State $L := 500$
\State

\textit{//  No. candidates in gallery.}
\State $N \gets len(I_G)$
\State

\State \textit{//  Initialize score lists.}
\State $S_1, S_2, S_3 := [0]*N, [0]*K, [0]*L$
\State

\State \textit{//  Extract gallery and probe features.}
\State $M_p, V_p, Z_p \gets Extract(I_p)$
\State $M_G, V_G, Z_G \gets Extract(I_G)$
\State

\State \textit{//  Stage 1 Matching}
\For{i in range(N)}
    \State $M_g, Z_g := M_G[i], Z_G[i]$
    \State $S_1[i] \gets w_{1}m_{simi}(M_p,M_g) + w_{2}\frac{(Z^{T}_{p}\cdot Z_{g})}{\lvert Z_{p}\rvert\lvert Z_{g}\rvert}$
\EndFor
\State $I^1_G, M^1_G, V^1_G, Z^1_G \gets SortAndFilterCandidates(S_1)$
\State

\State \textit{//  Stage 2 Matching}
\For{i in range(K)}
    \State $M_g, V_g, Z_g := M^1_G[i], V^1_G[i], Z^1_G[i]$
    \State $S_2[i] \gets w_{1}m_{simi}(M_p,M_g) + w_{2}\frac{(Z^{T}_{p}\cdot Z_{g})}{\lvert Z_{p}\rvert\lvert Z_{g}\rvert}$
        \State $      + w_{3}m_{simi}(V_p,V_g)$
\EndFor
\State $I^2_G, M^2_G, V^2_G, Z^2_G \gets SortAndFilterCandidates(S_2)$
\State

\State \textit{//  Stage 3 Matching}
\For{i in range(L)}
    \State $M_g, I_g, Z_g, V_g := M^2_G[i], I^2_G[i], Z^2_G[i], V^2_G[i]$
    \State $Z'_p, Z'_g \gets Realign(I_p,M_p,I_g,M_g)$
    \State $S_3[i] \gets w_{1}m_{simi}(M_p,M_g) + w_{2}\frac{(Z^{T}_{p}\cdot Z_{g})}{\lvert Z_{p}\rvert\lvert Z_{g}\rvert}$
        \State $      + w_{3}m_{simi}(V_p,V_g) + w_4\frac{(Z'^{T}_{p}\cdot Z'_{g})}{\lvert Z'_{p}\rvert\lvert Z'_{g}\rvert}$
\EndFor
\State $I^3_G, M^3_G, V^3_G, Z^3_G \gets SortAndFilterCandidates(S_3)$

\State 
\State \textit{//  Return sorted candidate list.}
\State \textbf{return} $I^3_G$

\EndProcedure
\end{algorithmic}
\end{algorithm}


\section{Experimental Results}
In this section, we report the performance of our latent fingerprint recognition pipeline across multiple latent fingerprint datasets, as well as other plain, rolled, and contactless fingerprint datasets to demonstrate the generalizability of our representations. First, we give the details of the datasets used in this study, followed by the closed-set and open-set identification results for several latent datasets, as well as the authentication performance across a diverse set of fingerprint sensors (e.g., capacitive, optical, etc.) and fingerprint type (plain, rolled, contactless, etc.). Next, we benchmark the performance of our enhancement network compared to previous enhancement methods, both in terms of minutiae detection accuracy and authentication performance of Verifinger v12.3 on each of the enhanced image outputs. Finally, we conclude with a discussion on the speed and computational efficiency of our recognition pipeline and the trade-offs in speed and accuracy given our multi-stage search strategy.

\begin{table}
\centering
\caption{Fingerprint Datasets used in this study.}
\label{tab:datasets}
\begin{tabular}{lcc}
\noalign{\hrule height 1.5pt}
\textbf{Train Datasets} & \textbf{\# Fingers} & \textbf{\# images} \\
\noalign{\hrule height 1.0pt}
MSP Rolled$^\dagger$~\cite{yoon2015longitudinal} & 37,411 & 447,988\\
\hline
\leftcell{NIST SD 302 (N2N)$^\ddagger$~\cite{nist302}\\(plain and rolled prints)} & 1,600 & 20,008\\
\hline
MSU Self-Collection (plain prints) & 4,582 & 57,813\\
\hline
\noalign{\hrule height 1.0pt}
\textbf{Validation Datasets} & \textbf{\# Fingers} & \textbf{\# Images} \\
\noalign{\hrule height 1.0pt}
\leftcell{NIST SD 302 (N2N)$^\ddagger$~\cite{nist302}\\(plain and rolled prints)} & 200 & 2,528\\
\hline
MSP Latent$^\dagger$~\cite{yoon2015longitudinal} & 933 & 2,030\\
\hline
\noalign{\hrule height 1.0pt}
\textbf{Test Datasets} & \textbf{\# Fingers} & \textbf{\# Images} \\
\noalign{\hrule height 1.0pt}
NIST SD 14~\cite{sd14} & 2700 & 5,400\\
\hline
NIST SD 302 (N2N)$^\ddagger$~\cite{nist302} & 200 & 2,548\\
\hline
\leftcell{NIST SD 302 Latents\\(N2N Latents)~\cite{nist302}} & 1,019 & 3,793\\
\hline
IIIT-D MOLF~\cite{sankaran2011matching} & 1,000 & 12,400\\
\hline
MSP Latent~\cite{yoon2015longitudinal} & 933 & 1,866\\
\hline
NIST SD 27~\cite{sd27} & 258 & 516\\
\hline
\leftcell{PolyU Contactless 2D to\\Contact-based 2D Database~\cite{lin2018matching}} & 160 & 960\\
\hline
\leftcell{ZJU Finger Photo and\\Touch-based Database~\cite{grosz2021c2cl}} & 824 & 19,776\\
\noalign{\hrule height 1.5pt}
\multicolumn{3}{p{0.95\linewidth}}{$^\dagger$ The MSP Rolled and MSP Latent datasets are completely disjoint and distinct in terms of finger identities.}\\
\multicolumn{3}{p{0.95\linewidth}}{$^\ddagger$ The train, validation, and test splits of N2N are disjoint and distinct in terms of finger identities.}\\
\end{tabular}
\vspace{-1.5em}
\end{table}

\subsection{Datasets}
Details for all training, validation, and test datasets used in this study are given in Table~\ref{tab:datasets}. Unlike previous latent fingerprint papers, we do not have access to a large private dataset of paired latent and rolled fingerprints (e.g., HiSign Latent Fingerprint database used in~\cite{gu2020latent, yang2014localized}, consisting of 10,458 latent and mated rolled pairs). In fact, we do not use any latent fingerprint datasets for training, yet our system is able to achieve new SOTA accuracy on many latent test datasets. Since our system is not highly tuned for latent fingerprints, we are able to maintain SOTA accuracy on rolled, plain, and contactless fingerprints as well.

\begin{table*}
\caption{Closed-set Identification (1:N comparison) results of three matchers, including the proposed LFR-Net.}
\label{tab:closed_set}
\begin{tabular}{c|c|c|c|c|c|c|c|c}
\noalign{\hrule height 1.5pt}
\multirow{3}{*}{\textbf{Model}} & \multirow{3}{*}{\specialcell{\textbf{Feature}\\\textbf{Extraction}\\\textbf{latency}$^\ddagger$}} & \multirow{3}{*}{\specialcell{\textbf{Comparison }\\\textbf{latency}$^\dagger$}} & \multirow{3}{*}{\specialcell{\textbf{Template size}\\\textbf{(latent/rolled)}}} & \multicolumn{5}{c}{\textbf{Closed-Set Rank 1 Retrieval Rate (\%)}}                                                                                      \\
\cline{5-9}
                                &                                                                                   &                                                                              &                                                              & \multirow{2}{*}{{\specialcell{\textbf{NIST SD 27}}}} & \multirow{2}{*}{{\specialcell{\textbf{N2N Latent}}}} & \multirow{2}{*}{{\specialcell{\textbf{MSP Latent}}}}  & \multicolumn{2}{c}{\textbf{MOLF}}\\
                                &                                                                                   &                                                                              &                                                              &                                      &                                                                    &    & \multicolumn{1}{c}{\textbf{DB1/DB4}} & \textbf{DB2/DB4}\\
\noalign{\hrule height 1.0pt}
\specialcell{MSU-AFIS~\cite{cao2019end}}                    & 2,586ms                                                                           & 0.093ms                                                                      & 308KB / 56KB                                                 & 61.63                              & 29.78                       & 67.64                                                                     & 44.84 & 27.86\\
\specialcell{AFR-Net~\cite{grosz2022afr}}                         & 6.86ms                                                                            & 0.002ms                                                                      & 3KB / 3KB                                                    & 39.92                              & 21.95                       & 59.27                                                                       & 42.41 & 38.48 \\
\specialcell{LFR-Net (proposed)}                         & 553ms                                                                             & 0.068ms*                                                                     & 307KB / 401KB                                                & \textbf{84.11}                              & \textbf{54.36}                       & \textbf{84.35}        & \textbf{70.43} &     \textbf{62.86}                                           \\
\noalign{\hrule height 1.5pt}
\multicolumn{9}{p{0.95\linewidth}}{$^\ddagger$ Computed on an Nvidia RTX A6000 GPU.}\\
\multicolumn{9}{p{0.95\linewidth}}{$^\dagger$ 128 threads on an AMD EPYC 7543 32-Core Processor.}\\
\multicolumn{9}{p{0.95\linewidth}}{$*$ Time computed using 3-stage matching with K=1,000 and L=500. Recall, K is the no. of candidates from the gallery passed from stage 1 to stage 2 and L is the no. of candidates passed from stage 2 to stage 3.}\\
\end{tabular}
\end{table*}

\begin{table}
\centering
\caption{Rank-1 retrieval rate (\%) of the proposed LFR-Net on NIST SD 27 gallery size ranging from 50,000 to 250,000.}
\label{tab:scale}
\begin{tabular}{c|c|c|c|c|c}
\noalign{\hrule height 1.5pt}
\specialcell{\textbf{Gallery Size (N)}} & \textbf{50K} & \textbf{100K} & \textbf{150K} & \textbf{200K} & \textbf{250K}\\
\noalign{\hrule height 1.0pt}
\specialcell{Rank-1 Retrieval Rate (\%)} & 85.66 & 84.11 & 83.33 & 82.56 & 82.17\\
\noalign{\hrule height 1.5pt}
\end{tabular}
\end{table}

\begin{table}
\centering
\caption{Open-set comparison of FNIR at FPIR=0.02 across 5 latent dataset evaluations (lower is better).}
\label{tab:open_set}
\begin{tabular}{c|c|c|c|c|c}
\noalign{\hrule height 1.5pt}
\multirow{2}{*}{{\specialcell{\textbf{Method}}}} & \multirow{2}{*}{{\specialcell{\textbf{NIST}\\\textbf{SD 27}}}} & \multirow{2}{*}{{\specialcell{\textbf{N2N}\\\textbf{Latent}}}} & \multirow{2}{*}{{\specialcell{\textbf{MSP}\\\textbf{Latent}}}}  & \multicolumn{2}{c}{\textbf{MOLF}}\\
 &  &  &  & \multicolumn{1}{c}{\textbf{DB1/DB4}} & \textbf{DB2/DB4}\\
\noalign{\hrule height 1.0pt}
\specialcell{MSU-AFIS\\~\cite{cao2019end}} & 0.72 & 0.86 & 0.69	& 0.76 & 0.87\\
\hline
\specialcell{LFR-Net\\(proposed)} & \textbf{0.50} & \textbf{0.74} & \textbf{0.44} & \textbf{0.60} & \textbf{0.68}\\
\noalign{\hrule height 1.5pt}
\end{tabular}
\end{table}

\begin{figure}
\includegraphics[width=\linewidth]{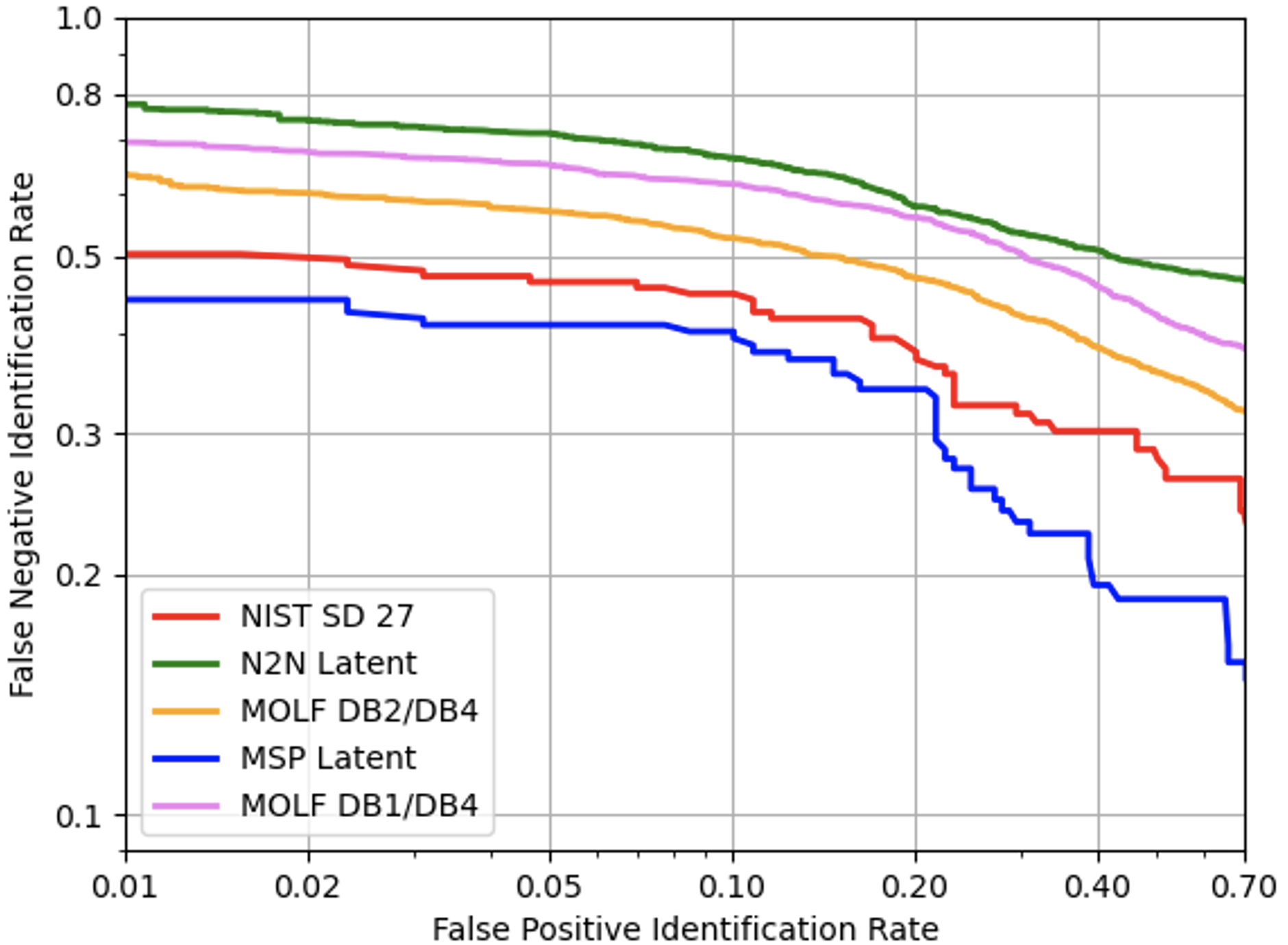} 
\caption{Open-set performance of LFR-Net with a gallery of 100K. Only 50\% of the latent probes in each dataset have mates in the gallery.}
\label{fig:open_set}
\end{figure}

\subsection{Identification Results}
Particularly relevant to the use case of latent fingerprint recognition is the ability to quickly and accurately search a large database of known fingerprints given an input probe latent fingerprint image. To benchmark the performance of our proposed pipeline for latent to rolled fingerprint recognition, we compute the closed-set and open-set identification results across four different latent fingerprint datasets against a gallery of 100K rolled fingerprints. Each fingerprint image in the gallery is from a unique finger and is separate from the rolled mates included in each latent database. Furthermore, all gallery fingerprints used in these experiments are from a separate set of the MSP forensic database~\cite{yoon2015longitudinal} and were not used in training.

A comparison of the rank-1 retrieval rate of the proposed method and two baseline algorithms for latent fingerprint recognition is shown in Table~\ref{tab:closed_set}. The two baseline methods include i.) the original AFR-Net as proposed in \cite{grosz2022afr} and ii.) an optimized version of the MSU-AFIS latent recognition pipeline proposed in \cite{cao2019end}. Currently, none of the commercial latent fingerprint vendors provide their latent fingerprint SDK to include in the evaluation. It is reported that Verifinger SDK v13 may have a latent matcher, but it is not yet available to developers at this time\footnote{\url{https://neurotechnology.com/verifinger.html}}. Additionally, many other previous latent fingerprint algorithms in the literature either have not made the source code publicly available or take prohibitively long to run the evaluation on our large size of gallery. Where available, we have included the performance of these baseline methods using the numbers reported in those publications in Table~\ref{tab:prior_work}. Not surprisingly, AFR-Net underperforms across each of the latent datasets compared to both MSU-AFIS and our proposed pipeline. MSU-AFIS performs reasonably well across each dataset because of its use of minutiae and virtual minutiae, which according to our ablation table in section~\ref{sec:ablation_study} makes a significant difference in the accuracy for latent to rolled comparisons. Nonetheless, our method, LFR-Net, outperforms all baseline methods due to a combination of improved enhancement, segmentation, and fusion of both local and global embeddings. In particular, our average rank-1 retrieval rate across the four datasets is 71.22\%, compared to the average rank-1 performance of MSU-AFIS of 46.19\%. Out of the published results on NIST SD 27, LFR-Net outperforms the next best method of Gu et al. 84.11\% to 70.1\% at rank-1.

For time and space constraints, most of our experiments are conducted on a background gallery of 100,000 unique rolled fingerprints; however, in order to investigate the scalability of LFR-Net to larger gallery sizes, we conducted closed-set identification experiments using NIST SD 27 on gallery sizes ranging from N=50,000 to N=250,000 rolled distractors. The results given in Table~\ref{tab:scale} suggest that the decline in rank-1 retrieval rate up to a gallery size of 250,000 starts to converge toward 82\%, down from an initial 85.66\% for gallery size of 50,000. As part of future work, we will investigate this trend up toward a gallery size of 1,000,000 unique fingers, which is more indicative of practical real-world applications.

To the best of our knowledge, previous studies on latent to rolled fingerprint matching have only reported closed-set identification results. With our improved performance across many of the latent to rolled datasets, we also report open-set identification results where $50\%$ of the probes from each of our datasets are randomly selected to have no corresponding mates in the gallery. A plot of false negative identification rate (FNIR) vs. false positive identification rate (FPIR), computed for rank-1 retrieval, for LFR-Net across all five latent evaluations is given in Figure~\ref{fig:open_set}, where a comparison of FNIR @ FPIR=0.02 with MSU-AFIS is given in Table~\ref{tab:open_set}.

\begin{table*}
\caption{Authentication (1:1 comparison) results of three matchers, including the proposed LFR-Net.}
\label{tab:authentication}
\begin{tabular}{c|ccccc|ccccc}
\noalign{\hrule height 1.5pt}
\multirow{2}{*}{\specialcell{\textbf{Model}}} & \multicolumn{5}{c|}{\textbf{TAR (\%) @ FAR=0.01\%}} & \multicolumn{5}{c}{\textbf{EER (\%)}}\\
\cline{2-11}
& \specialcell{\textbf{NIST SD 14}} & \specialcell{\textbf{NIST}\\\textbf{SD 302}} & \specialcell{\textbf{PolyU$^\ddagger$}} & \specialcell{\textbf{ZJU$^\ddagger$}} & \specialcell{\textbf{NIST SD 27$^\dagger$}} & \specialcell{\textbf{NIST SD 14}} & \specialcell{\textbf{NIST}\\\textbf{SD 302}} & \textbf{PolyU$^\ddagger$} & \textbf{ZJU$^\ddagger$} & \specialcell{\textbf{NIST SD 27$^\dagger$}}\\
\noalign{\hrule height 1.0pt}
\specialcell{Verifinger v12.3}                                                         &  99.93              & 93.26                                               & 95.39                  & 96.88       &  55.04 & \textbf{0.04} & 2.52 & 1.01 & 0.83 &   12.91   \\
AFR-Net~\cite{grosz2022afr}                           & 99.93              & 95.42                                     & 98.04                  & 98.78    & 59.69 & \textbf{0.04} & 2.03 & \textbf{0.34} & 0.50 &     11.52   \\
LFR-Net (stage 1) & \textbf{99.96} & \textbf{95.56}        &     \textbf{98.61}                  &     \textbf{99.00}           & \textbf{68.99}       & \textbf{0.04} & 1.96 & 0.45 & \textbf{0.45} &  7.74          \\
LFR-Net (stage 1\&2) &     99.93  &      94.30        &     \textbf{98.61}                  &     \textbf{99.00}           & \textbf{68.99}               & \textbf{0.04} & \textbf{1.87} & 0.45 & \textbf{0.45} &  \textbf{6.58}          \\
\noalign{\hrule height 1.5pt}
\multicolumn{11}{p{0.95\linewidth}}{$^\dagger$ Using LFR-Net preprocessing and segmentation.}\\
\multicolumn{11}{p{0.95\linewidth}}{$^\ddagger$ Contactless images are preprocessed using the enhancement and unwarping method proposed in \cite{grosz2021c2cl}}.\\
\end{tabular}
\end{table*}

\subsection{Authentication Performance}
Due to the challenges inherent to latent fingerprint recognition, all the existing systems in the open literature are highly tuned to perform well for latent to rolled comparison and/or require expensive feature extraction and matching times due to the additional features required to achieve high accuracy. However, our system stands out in that the representations learned (both local and global embeddings) are generalizable across a wide range of fingerprint image characteristics, as demonstrated in the authentication results shown in Table~\ref{tab:authentication}. LFR-Net is competitive with and even outperforms the commercial fingerprint SDK Verifinger v12.3 on several datasets.

Furthermore, our algorithm can be tuned to vary the latency for both feature extraction and matching depending on the confidence required and/or difficulty of the fingerprint image domain of interest. For example, both the feature extraction and matching latencies could be significantly decreased for rolled-to-rolled or plain-to-plain fingerprint matching by utilizing just the AFR-Net embeddings, which achieves very competitive authentication performance across all of the datasets. Similarly, at a modest cost in latency, one could also incorporate minutiae features for a slight boost in accuracy, as is done in stage 1 of LFR-Net. Virtual minutiae are effective in improving latent to rolled matching accuracy; however, they introduce additional latency which is not required to achieve high accuracy across rolled and plain fingerprint datasets.

\subsection{Latent Enhancement Performance}
For a comparison of our enhancement method with several previous SOTA latent enhancement methods (FingerGAN~\cite{zhu2023fingergan}, FingerNet~\cite{tang2017fingernet}, and MSU-AFIS~\cite{cao2019end}) we computed the statistics of false (spurious) and correctly predicted minutiae using the manually marked ground truth provided for NIST SD 27 from \cite{feng2012orientation}. Specifically, we individually enhanced all 258 latent images from NIST SD 27 using each of the enhancement methods, extracted the minutiae points of the enhanced images using Verifinger v12.3, and compared the extracted minutiae to the human annotated ground truth minutiae points. We consider a correctly detected minutiae as one in which the type is the same, (x,y) location is within 10 pixels, and the angle difference is less than 10 degrees compared to a ground truth minutiae. These thresholds are motivated from a previous study on the robustness of minutiae-based matchers, which showed that the performance of minutiae matching starts to decline with minutiae perturbations outside these ranges~\cite{grosz2020white}. Results in Table~\ref{tab:minu_stats} show that our enhancement network outperforms the previous methods in terms of number of correctly predicted minutiae, while also introducing fewer spurious minutiae than the next best method, FingerGAN. Interestingly, the number of spurious predicted minutiae is the lowest for the unenhanced, original latent images. The increase in spurious minutiae may, at least partially, be attributed to each enhancement method hallucinating fingerprint ridges where there are none; however, the number of spurious minutiae increasing dramatically across all methods suggests that the increase could be attributed in some part to missed minutiae during the manual markup process due to the difficulty in annotating minutiae for unenhanced latent fingerprints. Nonetheless, according to previous research, spurious minutiae have much less of an effect on overall matching accuracy as does failing to detect correct minutiae~\cite{grosz2020white}. In fact, we can see this is indeed the case in the improved authentication (1:1 matching) performance of Verifinger on our enhanced NIST SD 27 images (TAR=65.12\% at FAR=0.1\%) compared to the original images (TAR=57.75\% at FAR=0.1\%), where a total of 258 genuine scores and 66,564 imposter scores were computed.

Interestingly, the rank-1 performance using Verifinger is the same for our enhanced images as those of FingerGAN~\cite{zhu2023fingergan}; however, we have the additional advantage that our enhanced images are not introducing a domain shift with respect to the original, gray-scale fingerprint images. This means that other components of our pipeline (e.g., minutiae descriptor and AFR-Net) can also benefit from the enhancement without the need to be re-trained. For example, the rank-1 performance of AFR-Net on NIST SD 27 enhanced by LFR-Net improves from 39.92\% to 52.33\% without re-training the network, but the performance drops to 6.20\% using FingerGAN enhanced images.

\begin{table*}
\centering
\caption{Number of correctly predicted minutiae and spurious minutiae introduced for the 258 latent prints in NIST SD 27 before and after enhancement by several methods. Predicted minutiae were extracted using Verifinger v12.3 and ground truth minutiae were manually marked by \cite{feng2012orientation}.}
\label{tab:minu_stats}
\begin{tabular}{c|c|c|c|c}
\noalign{\hrule height 1.5pt}
\specialcell{\textbf{Method}} & \specialcell{\textbf{\# Correctly Predicted}} & \specialcell{\textbf{\# Spurious}} & \specialcell{\textbf{TAR (\%) @ FAR=0.01\% (0.1\%)}} & \specialcell{\textbf{Rank-1 (\%) with gallery of 100K}}\\
\noalign{\hrule height 1.0pt}
\specialcell{Original Images} & 2,606 & \textbf{3,676} & 51.94 (57.75) & 56.59 \\
\hline
\specialcell{Enhanced by MSU-AFIS~\cite{cao2019end}} & 2,329 & 4,678 & 38.76 (44.96) & 48.06\\
\hline
\specialcell{Enhanced by FingerNet~\cite{tang2017fingernet}} & 2,460 & 5,147 & 39.15 (47.29) & 47.29\\
\hline
\specialcell{Enhanced by FingerGAN~\cite{zhu2023fingergan}} & 2,939 & 7,385 & 52.71 (57.36) & \textbf{58.14} \\
\hline
\specialcell{Enhanced by LFR-Net \\(Proposed)} & \textbf{3,118} & 5,536 & \textbf{55.04 (65.12)} & \textbf{58.14} \\
\noalign{\hrule height 1.5pt}
\end{tabular}
\end{table*}

\subsection{Computational Efficiency}
\label{sec:latency}
Latency is a crucial aspect for large-scale identification applications, which tends to be in competition with accuracy. Thus, we were motivated to find a balance between accuracy and speed using a multi-stage search protocol, which has also been explored in previous works on fingerprint identification~\cite{deepprint}. For a quantitative analysis on the latency of our approach, we will denote the size of our gallery as N (e.g., N=100,000) and the size of our probe dataset as Q (e.g., Q=258 in the case of NIST SD 27). Furthermore, since our algorithm consists of three stages of matching with variable number of top candidates per probe passed to subsequent stages, we will denote the number of candidates per probe image passed from the first stage to the second stage as K and the number of candidates passed to the third stage as L.

For our first stage matching, we utilize only AFR-Net and minutiae features to obtain a short list of top K candidates from the gallery for each probe fingerprint image. This stage takes on average $t_1$=0.015 ms for a single latent to rolled comparison when utilizing 128 threads on an AMD EPYC 7543 32-Core Processor, where a total of NxQ comparisons are computed. In the second stage, we utilize virtual minutiae scores to re-rank the K list of candidates per latent and return a further condensed list of top L candidates to pass to the third stage. Here, a single virtual minutiae comparison between a latent and rolled image pair takes on average $t_2$=0.984 ms, where a total of KxQ comparisons are computed. Finally, our third stage consists of re-aligning each of the L candidate images for each probe using the pairwise minutiae correspondences and recomputing AFR-Net scores for each pair. In this stage, there are a total of LxQ comparisons required, where each realignment plus AFR-Net inference per comparison takes an average of $t_3$=8.626 ms. Note, the latency of stage 1 and stage 2 depends on the number of minutiae and virtual minutiae extracted per latent probe, respectively. The latency numbers reported here are computed for NIST SD 27 against a gallery augmented by 100,000 rolled fingerprints, where the average number of minutiae and virtual minutiae extracted per latent image is 45 and 363, respectively, and the average number of minutiae and virtual minutiae per rolled fingerprint is 119 and 886, respectively. In total, the average latency $t$ per comparison for the entire three stage matching process can be computed using equation~\ref{eq:latency}:

\begin{equation}
\label{eq:latency}
    t = t_1 + \frac{K}{N}t_2 + \frac{L}{N}t_3
\end{equation}

\begin{figure}
\includegraphics[width=\linewidth]{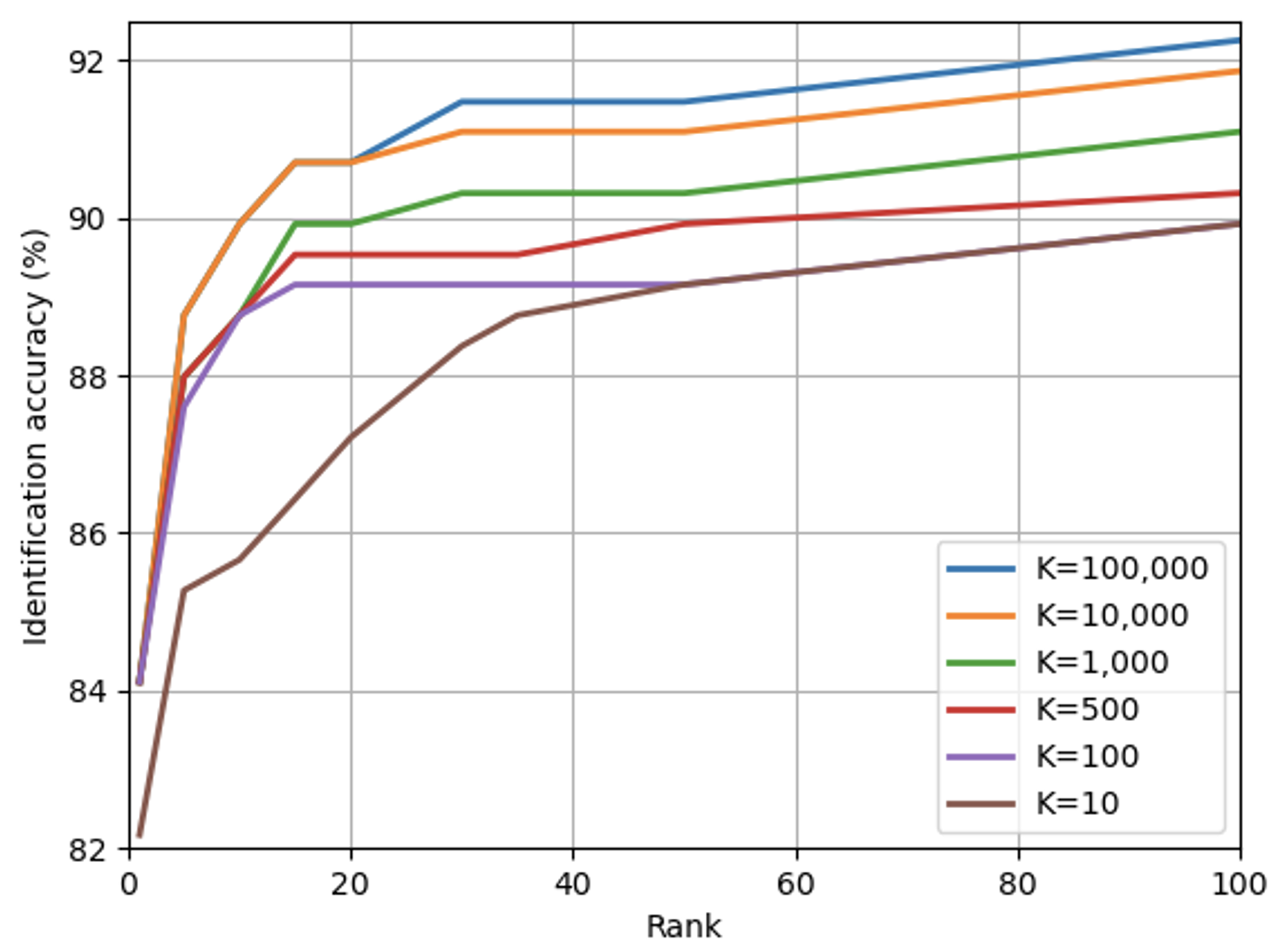} 
\caption{Search results on NIST SD 27 (100K gallery) as the no. of candidates (K) sent to stage 2 is varied. Reducing K to 1\% of the gallery results in a speed up of 0.352 ms to 0.068 ms (5.2$\times$ faster) per comparison with no change in rank-1 accuracy and only $\sim1$\% decrease at higher ranks.}
\label{fig:cmc}
\end{figure}

Using equation~\ref{eq:latency} with N=100,000, K=1000 and L=500, the average latent to rolled comparison across each of the four latent datasets for our full matching pipeline takes about t=0.068 ms. As mentioned previously, the filtering of our candidate lists in each stage does incur some accuracy trade-off; however, we find that filtering $99\%$ of the candidate list prior to stage 2 (with K=1,000 and N=100,000) leads to no difference in rank-1 retrieval rate for NIST SD 27 and only about a 1\% decrease in accuracy at higher ranks. A plot of the Cumulative Match Characteristic (CMC) for NIST SD 27 on a gallery of 100,000 as the value of K is varied from 100,000 to 10 is shown in Figure~\ref{fig:cmc}.

The feature extraction speed is often less of a concern for fingerprint recognition since templates for the gallery can be extracted offline prior to matching; however, is still important in cases of updating the gallery for future improvements to the system. Nonetheless, our method is significantly faster compared to the baseline MSU-AFIS algorithm, taking just 553 ms on average per latent image or 1.88 images per second. In terms of template size, our algorithm is comparable to MSU-AFIS for latents; however, for rolled templates, MSU-AFIS performs several template compression and quantization techniques to reduce the size of the templates compared to ours, which can also be incorporated into our algorithm in future work.

\section{Discussion}
In this section we discuss some of the failure cases of our pipeline and present plausible strategies to improve on these challenging cases. Furthermore, we present an ablation analysis to evaluate the contribution of each component of our pipeline to the overall identification accuracy.

\subsection{Failure Case Analysis}
\begin{figure}
\includegraphics[width=\linewidth]{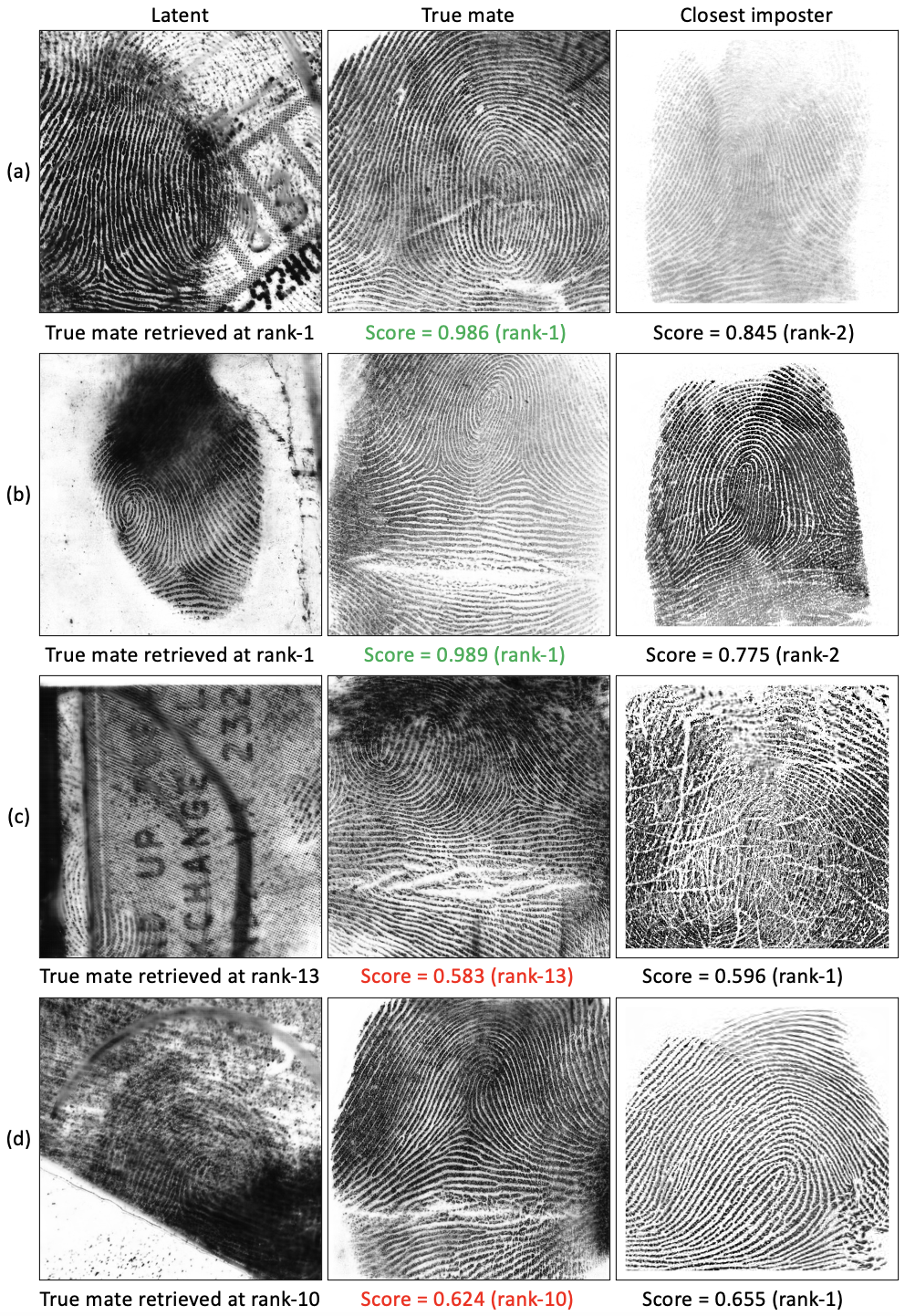} 
\caption{Example success (a and b) and failure (c and d) cases of the proposed LFR-Net on the NIST SD 27 latent database.}
\label{fig:failures_and_successes}
\vspace{-1.5em}
\end{figure}

Despite the SOTA performance of our algorithm across all five latent evaluation fingerprint datasets utilized, there are still cases in which our algorithm fails to return the correct mate at rank one (see Figure~\ref{fig:failures_and_successes} (c) and (d)). Figure~\ref{fig:failures_and_successes} (a) and (b) show examples where the system was successful in returning the correct mate at rank-1, demonstrating the benefit of the enhancement network in removing some occlusions and enhancing the inter-ridge separation. However, as examples (c) and (d) show, there are many challenges that need to be tackled. One cause of failure, demonstrated in (c), is poor segmentation, where a large portion of the fingerprint image is cut-off. Additional failures can be attributed to noisy background and overlapping latent patterns, very low ridge-valley contrast, and extreme rotations. In case of small area latent fingerprints, it becomes difficult to estimate the correct rotation to align latent fingerprints prior to global feature extraction. A simple way to overcome this difficulty could be to rotate the latent image 4 times by $90\degree$ increments and take a max fusion of the global similarity scores obtained when matching each of the four rotated images with all the images in the gallery. Perhaps an even better approach would be to make the global embeddings inherently rotation invariant, which will be one focus of our future work.

\subsection{Ablation Study}
\label{sec:ablation_study}
To justify the use of each component (enhancement, minutiae, virtual minutiae, global embedding, and realignment stage), we perform an ablation analysis as each additional module is added to the matching pipeline. The results are given in Table~\ref{tab:ablation}, which show improved search performance with the addition of each module. We observe a significant jump in accuracy with the incorporation of local minutiae features, another significant jump in accuracy with virtual minutiae, and a final improvement in using the realignment stage. The performance across all datasets starts to saturate after the second stage matching with virtual minutiae, where the third stage (realignment) adds the most noticeable benefit for datasets with extreme rotations (e.g., N2N Latent). We also see significant improvements in rank-1 retrieval with using our enhancement network vs. without any enhancement. For example, the performance on NIST SD 27 improves from $72.87\%$ without enhancement to $84.11\%$ with enhancement.

\begin{table*}
\centering
\caption{LFR-Net ablation study.}
\label{tab:ablation}
\begin{tabular}{c|c|c|c|c||c|c|c|c|c}
\noalign{\hrule height 1.5pt}
\multicolumn{5}{c||}{\textbf{Modules}} & \multicolumn{5}{c}{\textbf{Rank-1 accuracy (\%) on a gallery of 100K}}\\
\noalign{\hrule height 1.0pt}
\specialcell{\textbf{Global}\\\textbf{Embedding}} & \textbf{Minutiae} & \specialcell{\textbf{Virtual}\\\textbf{Minutiae}} & \textbf{Realign} & \textbf{Enhance} & \specialcell{\textbf{NIST SD 27}} & \specialcell{\textbf{N2N Latent}} & \specialcell{\textbf{MSP Latent}} & \specialcell{\textbf{MOLF}\\\textbf{(DB1/DB4)}} & \specialcell{\textbf{MOLF}\\\textbf{(DB2/DB4)}}\\
\noalign{\hrule height 1.0pt}
\checkmark & & & & & 39.92 & 21.95 & 59.27 & 42.41 & 38.48 \\
\hline
 & \checkmark & & & & 57.75 & 46.47 & 72.45 & 45.05 & 33.36 \\
\hline
\checkmark & \checkmark & & & & 65.12 & 46.58 & 78.67 & 54.86 & 45.20 \\
\hline
\checkmark & \checkmark & \checkmark & & & 72.48 & 49.21 & 80.71 & 60.34 & 50.25\\
\hline
\checkmark & \checkmark & \checkmark & \checkmark &  & 72.87 & 50.25 & 81.78 & 60.77 & 50.52 \\
\hline
\checkmark &  &  &  & \checkmark & 52.33 & 23.00 & 60.34 & 53.48 & 49.85\\
\hline
 & \checkmark &  &  & \checkmark & 67.69 & 51.66 & 78.24 & 60.52 & 50.32\\
\hline
\checkmark & \checkmark &  &  & \checkmark & 75.58 & 51.12 & 81.46 & 67.18 & 59.68\\
\hline
\checkmark & \checkmark & \checkmark &  & \checkmark & \textbf{84.11} & 53.50 & 83.92 & 70.14 & 62.68\\
\hline
\checkmark & \checkmark & \checkmark & \checkmark & \checkmark & \textbf{84.11} & \textbf{54.36} & \textbf{84.35} & \textbf{70.43} & \textbf{62.86}\\
\noalign{\hrule height 1.5pt}
\end{tabular}
\end{table*}

\section{Conclusion}
In this paper, we presented a pipeline for end-to-end latent fingerprint recognition and demonstrated its SOTA performance across five different latent fingerprint evaluations (for both closed-set and open-set identification), as well as its generalization across several rolled, plain, and contact to contactless fingerprint datasets. Our network incorporates a novel use of both local (minutiae and virtual minutiae) and global (AFR-Net) embeddings for improved latent fingerprint recognition. We also present a multi-stage search strategy to decrease the time required for large-scale identification, which is adaptable for a desired trade-off in accuracy and search speed.

\section{Acknowledgment}
This research was supported by a grant from the Department of Homeland Security via The Criminal Investigations and Network Analysis Center (CINA) at George Mason University.

\ifCLASSOPTIONcaptionsoff
  \newpage
\fi

\bibliography{cite}

\begin{thebibliography}{10}

\bibitem{nist302}
G.~P. Fiumara, P.~A. Flanagan, J.~D. Grantham, K.~Ko, K.~Marshall, M.~Schwarz,
  E.~Tabassi, B.~Woodgate, and C.~Boehnen, ``Nist special database 302: Nail to
  nail fingerprint challenge,'' Tech. Rep. NIST.TN.2007, National Institute of
  Standards and Technology, Gaithersburg, MD, 2019.

\bibitem{yoon2015longitudinal}
S.~Yoon and A.~K. Jain, ``Longitudinal study of fingerprint recognition,'' {\em
  Proceedings of the National Academy of Sciences}, vol.~112, no.~28,
  pp.~8555--8560, 2015.

\bibitem{sd27}
M.~D. Garris, {\em NIST special database 27: Fingerprint minutiae from latent
  and matching tenprint images}.
\newblock US Department of Commerce, National Institute of Standards and
  Technology, 2000.

\bibitem{sankaran2015multisensor}
A.~Sankaran, M.~Vatsa, and R.~Singh, ``Multisensor optical and latent
  fingerprint database,'' {\em IEEE access}, vol.~3, pp.~653--665, 2015.

\bibitem{yamashita2010fingerprint}
B.~Yamashita and M.~French, ``Fingerprint sourcebook-chapter 7: Latent print
  development,'' {\em US Dept. of Justice, Office of Justice Programs, National
  Institute of Justice}, 2010.

\bibitem{handbook}
D.~Maltoni, D.~Maio, A.~K. Jain, and F.~Jianjiang, {\em Handbook of
  {F}ingerprint {R}ecognition}.
\newblock Springer Science \& Business Media, 3rd~ed., 2022.

\bibitem{oig2006review}
A.~Oig, ``Review of the fbi’s handling of the brandon mayfield case,'' {\em
  Office of the Inspector General, Oversight and Review Division, US Department
  of Justice}, pp.~1--330, 2006.

\bibitem{ashbaugh1999quantitative}
D.~R. Ashbaugh, {\em Quantitative-qualitative friction ridge analysis: an
  introduction to basic and advanced ridgeology}.
\newblock CRC press, 1999.

\bibitem{cao2014segmentation}
K.~Cao, E.~Liu, and A.~K. Jain, ``Segmentation and enhancement of latent
  fingerprints: A coarse to fine ridgestructure dictionary,'' {\em IEEE
  transactions on pattern analysis and machine intelligence}, vol.~36, no.~9,
  pp.~1847--1859, 2014.

\bibitem{li2018deep}
J.~Li, J.~Feng, and C.-C.~J. Kuo, ``Deep convolutional neural network for
  latent fingerprint enhancement,'' {\em Signal Processing: Image
  Communication}, vol.~60, pp.~52--63, 2018.

\bibitem{huang2020latent}
X.~Huang, P.~Qian, and M.~Liu, ``Latent fingerprint image enhancement based on
  progressive generative adversarial network,'' in {\em Proceedings of the
  IEEE/CVF Conference on Computer Vision and Pattern Recognition Workshops},
  pp.~800--801, 2020.

\bibitem{joshi2019latent}
I.~Joshi, A.~Anand, M.~Vatsa, R.~Singh, S.~D. Roy, and P.~Kalra, ``Latent
  fingerprint enhancement using generative adversarial networks,'' in {\em 2019
  IEEE winter conference on applications of computer vision (WACV)},
  pp.~895--903, IEEE, 2019.

\bibitem{zhu2023fingergan}
Y.~Zhu, X.~Yin, and J.~Hu, ``Fingergan: A constrained fingerprint generation
  scheme for latent fingerprint enhancement,'' {\em IEEE Transactions on
  Pattern Analysis and Machine Intelligence}, 2023.

\bibitem{ozturk2022minnet}
H.~{\.I}. {\"O}zt{\"u}rk, B.~Selbes, and Y.~Artan, ``Minnet: Minutia patch
  embedding network for automated latent fingerprint recognition,'' in {\em
  Proceedings of the IEEE/CVF Conference on Computer Vision and Pattern
  Recognition}, pp.~1627--1635, 2022.

\bibitem{tang2017latent}
Y.~Tang, F.~Gao, and J.~Feng, ``Latent fingerprint minutia extraction using
  fully convolutional network,'' in {\em 2017 IEEE International Joint
  Conference on Biometrics (IJCB)}, pp.~117--123, IEEE, 2017.

\bibitem{darlow2017fingerprint}
L.~N. Darlow and B.~Rosman, ``Fingerprint minutiae extraction using deep
  learning,'' in {\em IEEE International Joint Conference on Biometrics},
  pp.~22--30, IEEE, 2017.

\bibitem{cao2018latent}
K.~Cao and A.~K. Jain, ``Latent fingerprint recognition: role of texture
  template,'' in {\em 2018 IEEE 9th international conference on biometrics
  theory, applications and systems (BTAS)}, pp.~1--9, IEEE, 2018.

\bibitem{yang2014localized}
X.~Yang, J.~Feng, and J.~Zhou, ``Localized dictionaries based orientation field
  estimation for latent fingerprints,'' {\em IEEE transactions on pattern
  analysis and machine intelligence}, vol.~36, no.~5, pp.~955--969, 2014.

\bibitem{feng2012orientation}
J.~Feng, J.~Zhou, and A.~K. Jain, ``Orientation field estimation for latent
  fingerprint enhancement,'' {\em IEEE transactions on pattern analysis and
  machine intelligence}, vol.~35, no.~4, pp.~925--940, 2012.

\bibitem{cao2019end}
K.~Cao, D.-L. Nguyen, C.~Tymoszek, and A.~K. Jain, ``End-to-end latent
  fingerprint search,'' {\em IEEE Transactions on Information Forensics and
  Security}, vol.~15, pp.~880--894, 2019.

\bibitem{cao2018automated}
K.~Cao and A.~K. Jain, ``Automated latent fingerprint recognition,'' {\em IEEE
  Transactions on Pattern Analysis and Machine Intelligence}, vol.~41, no.~4,
  pp.~788--800, 2018.

\bibitem{tang2017fingernet}
Y.~Tang, F.~Gao, J.~Feng, and Y.~Liu, ``Fingernet: An unified deep network for
  fingerprint minutiae extraction,'' in {\em IEEE International Joint
  Conference on Biometrics}, pp.~108--116, IEEE, 2017.

\bibitem{sd14}
C.~I. Watson, ``Nist special database 14,'' {\em Fingerprint Database, US
  National Institute of Standards and Technology}, 1993.

\bibitem{lin2018matching}
C.~Lin and A.~Kumar, ``Matching contactless and contact-based conventional
  fingerprint images for biometrics identification,'' {\em IEEE Transactions on
  Image Processing}, vol.~27, no.~4, pp.~2008--2021, 2018.

\bibitem{grosz2021c2cl}
S.~A. Grosz, J.~J. Engelsma, E.~Liu, and A.~K. Jain, ``C2cl: Contact to
  contactless fingerprint matching,'' {\em IEEE Transactions on Information
  Forensics and Security}, vol.~17, pp.~196--210, 2021.

\bibitem{gu2020latent}
S.~Gu, J.~Feng, J.~Lu, and J.~Zhou, ``Latent fingerprint registration via
  matching densely sampled points,'' {\em IEEE Transactions on Information
  Forensics and Security}, vol.~16, pp.~1231--1244, 2020.

\bibitem{grosz2022afr}
S.~A. Grosz and A.~K. Jain, ``Afr-net: Attention-driven fingerprint recognition
  network,'' {\em arXiv preprint arXiv:2211.13897}, 2022.

\bibitem{cappelli2009semi}
R.~Cappelli, D.~Maio, and D.~Maltoni, ``Semi-automatic enhancement of very low
  quality fingerprints,'' in {\em 2009 Proceedings of 6th International
  Symposium on Image and Signal Processing and Analysis}, pp.~678--683, IEEE,
  2009.

\bibitem{chikkerur2007fingerprint}
S.~Chikkerur, A.~N. Cartwright, and V.~Govindaraju, ``Fingerprint enhancement
  using stft analysis,'' {\em Pattern Recognition}, vol.~40, no.~1,
  pp.~198--211, 2007.

\bibitem{yoon2011latent}
S.~Yoon, J.~Feng, and A.~K. Jain, ``Latent fingerprint enhancement via robust
  orientation field estimation,'' in {\em 2011 international joint conference
  on biometrics (IJCB)}, pp.~1--8, IEEE, 2011.

\bibitem{zhang2012latent}
J.~Zhang, R.~Lai, and C.-C.~J. Kuo, ``Latent fingerprint segmentation with
  adaptive total variation model,'' in {\em 2012 5th IAPR International
  Conference on Biometrics (ICB)}, pp.~189--195, IEEE, 2012.

\bibitem{zhang2013adaptive}
J.~Zhang, R.~Lai, and C.-C.~J. Kuo, ``Adaptive directional total-variation
  model for latent fingerprint segmentation,'' {\em IEEE Transactions on
  Information Forensics and Security}, vol.~8, no.~8, pp.~1261--1273, 2013.

\bibitem{cao2015latent}
K.~Cao and A.~K. Jain, ``Latent orientation field estimation via convolutional
  neural network,'' in {\em 2015 International Conference on Biometrics (ICB)},
  pp.~349--356, IEEE, 2015.

\bibitem{svoboda2017generative}
J.~Svoboda, F.~Monti, and M.~M. Bronstein, ``Generative convolutional networks
  for latent fingerprint reconstruction,'' in {\em 2017 IEEE International
  Joint Conference on Biometrics (IJCB)}, pp.~429--436, IEEE, 2017.

\bibitem{liu2020automatic}
M.~Liu and P.~Qian, ``Automatic segmentation and enhancement of latent
  fingerprints using deep nested unets,'' {\em IEEE Transactions on Information
  Forensics and Security}, vol.~16, pp.~1709--1719, 2020.

\bibitem{dabouei2018id}
A.~Dabouei, H.~Kazemi, S.~M. Iranmanesh, J.~Dawson, N.~M. Nasrabadi, {\em
  et~al.}, ``Id preserving generative adversarial network for partial latent
  fingerprint reconstruction,'' in {\em 2018 IEEE 9th International Conference
  on Biometrics Theory, Applications and Systems (BTAS)}, pp.~1--10, IEEE,
  2018.

\bibitem{beheshti2020squeeze}
N.~Beheshti and L.~Johnsson, ``Squeeze u-net: A memory and energy efficient
  image segmentation network,'' in {\em Proceedings of the IEEE/CVF conference
  on computer vision and pattern recognition workshops}, pp.~364--365, 2020.

\bibitem{cappelli2010minutia}
R.~Cappelli, M.~Ferrara, and D.~Maltoni, ``Minutia cylinder-code: A new
  representation and matching technique for fingerprint recognition,'' {\em
  IEEE transactions on pattern analysis and machine intelligence}, vol.~32,
  no.~12, pp.~2128--2141, 2010.

\bibitem{gu2022latent}
S.~Gu, J.~Feng, J.~Lu, and J.~Zhou, ``Latent fingerprint indexing: Robust
  representation and adaptive candidate list,'' {\em IEEE Transactions on
  Information Forensics and Security}, vol.~17, pp.~908--923, 2022.

\bibitem{deng2019arcface}
J.~Deng, J.~Guo, N.~Xue, and S.~Zafeiriou, ``Arcface: Additive angular margin
  loss for deep face recognition,'' in {\em Proceedings of the IEEE/CVF
  conference on computer vision and pattern recognition}, pp.~4690--4699, 2019.

\bibitem{sd4}
C.~I. Watson and C.~L. Wilson, ``Nist special database 4,'' {\em Fingerprint
  Database, National Institute of Standards and Technology}, vol.~17, no.~77,
  p.~5, 1992.

\bibitem{engelsma2022printsgan}
J.~J. Engelsma, S.~A. Grosz, and A.~K. Jain, ``Printsgan: synthetic fingerprint
  generator,'' {\em IEEE Transactions on Pattern Analysis and Machine
  Intelligence}, 2022.

\bibitem{grosz2022spoofgan}
S.~A. Grosz and A.~K. Jain, ``Spoofgan: Synthetic fingerprint spoof images,''
  {\em IEEE Transactions on Information Forensics and Security}, 2022.

\bibitem{ericson2015evaluation}
L.~Ericson and S.~Shine, ``Evaluation of contactless versus contact fingerprint
  data phase 2 (version 1.1),'' Tech. Rep. 249552, DOJ Office Justice Programs,
  I. ManTech Adv. Syst. Int., Fairmont, WV, 2015.

\bibitem{deb2018matching}
D.~Deb, T.~Chugh, J.~Engelsma, K.~Cao, N.~Nain, J.~Kendall, and A.~K. Jain,
  ``Matching fingerphotos to slap fingerprint images,'' {\em arXiv preprint
  arXiv:1804.08122}, 2018.

\bibitem{birajadar2019towards}
P.~Birajadar, M.~Haria, P.~Kulkarni, S.~Gupta, P.~Joshi, B.~Singh, and
  V.~Gadre, ``Towards smartphone-based touchless fingerprint recognition,''
  {\em S{\=a}dhan{\=a}}, vol.~44, no.~7, pp.~1--15, 2019.

\bibitem{wyzykowski2022synthetic}
A.~B.~V. Wyzykowski and A.~K. Jain, ``Synthetic latent fingerprint generator,''
  {\em IEEE/CVF Winter Conference on Applications of Computer Vision}, 2022.

\bibitem{sankaran2011matching}
A.~Sankaran, T.~I. Dhamecha, M.~Vatsa, and R.~Singh, ``On matching latent to
  latent fingerprints,'' in {\em 2011 international joint conference on
  biometrics (IJCB)}, pp.~1--6, IEEE, 2011.

\bibitem{grosz2020white}
S.~A. Grosz, J.~J. Engelsma, N.~G. Paulter, and A.~K. Jain, ``White-box
  evaluation of fingerprint matchers: Robustness to minutiae perturbations,''
  in {\em 2020 IEEE International Joint Conference on Biometrics (IJCB)},
  pp.~1--10, IEEE, 2020.

\bibitem{deepprint}
J.~J. Engelsma, K.~Cao, and A.~K. Jain, ``Learning a fixed-length fingerprint
  representation,'' {\em IEEE Transactions on Pattern Analysis and Machine
  Intelligence}, 2019.

\end{thebibliography}
\bibliographystyle{ieeetr}

\begin{IEEEbiography}[{\includegraphics[width=1in,height=1.25in,clip,keepaspectratio]{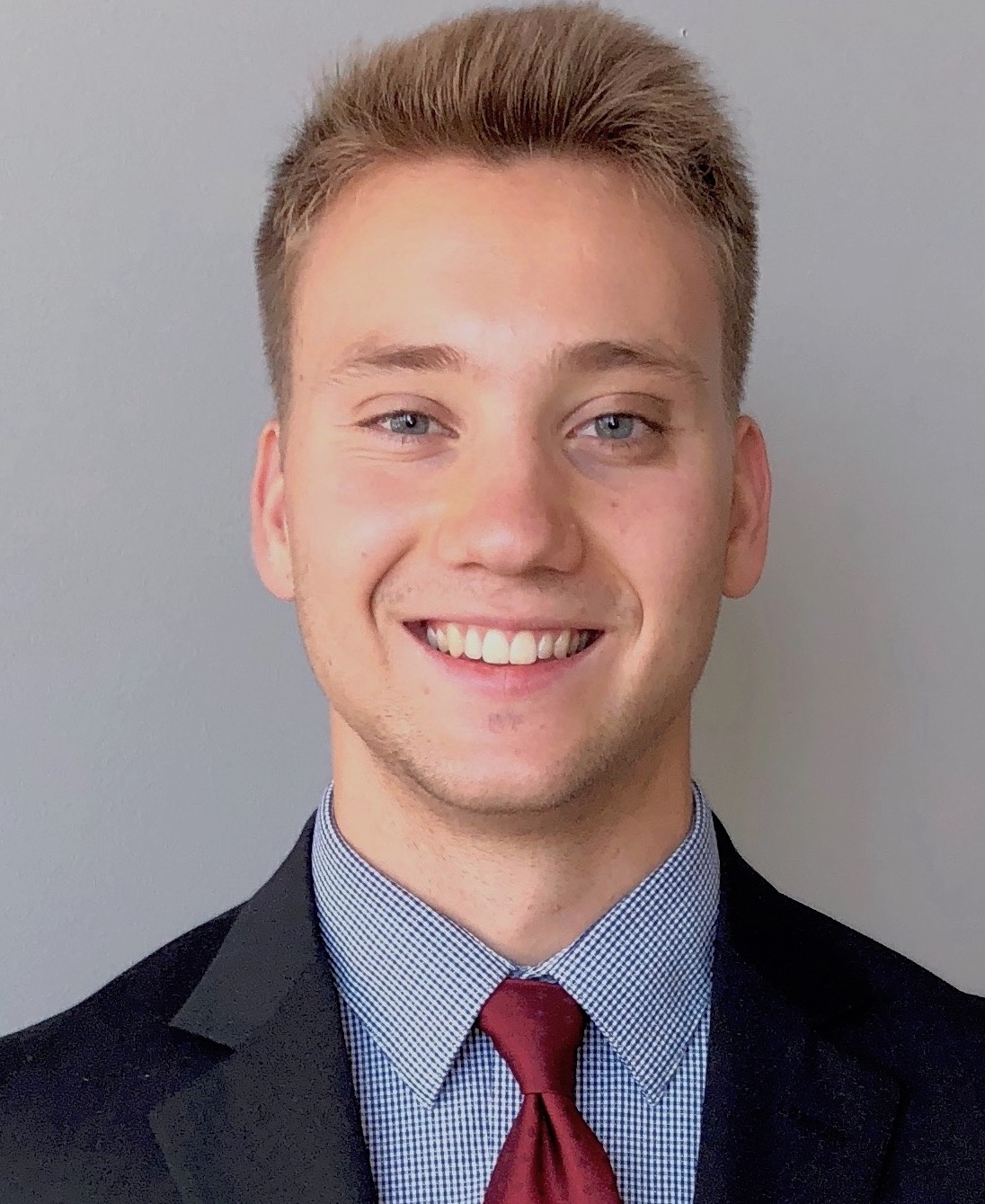}}]{Steven A. Grosz}
received his B.S. degree with highest honors in Electrical Engineering from Michigan State University, East Lansing, Michigan, in 2019. He is currently a doctoral student in the Department of Computer Science and Engineering at Michigan State University. His primary research interests are in the areas of machine learning and computer vision with applications in biometrics.
\vspace{-1.5em}
\end{IEEEbiography}

\begin{IEEEbiography}[{\includegraphics[width=1in,height=1.25in,clip,keepaspectratio]{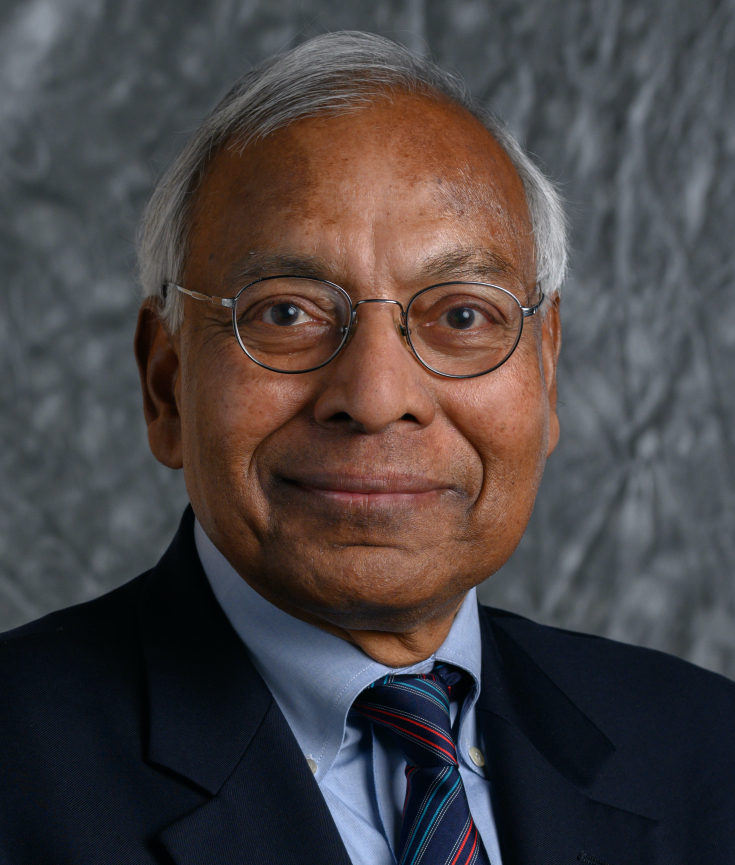}}]{Anil K. Jain}
Anil K. Jain is a University distinguished professor in the Department of Computer Science and Engineering at Michigan State University. His research interests include pattern recognition, computer vision, and biometric authentication. He served as the editor-in-chief of the IEEE Transactions on Pattern Analysis and Machine Intelligence and was a member of the United States Defense Science Board. He has received Fulbright, Guggenheim, Alexander von Humboldt, and IAPR King Sun Fu awards. He is a member of the National Academy of Engineering, the Indian National Academy of Engineering, the World Academy of Sciences, and the Chinese Academy of Sciences.
\vspace{-1.5em}
\end{IEEEbiography}

\end{document}